\pdfoutput=1
\documentclass{llncs}
\PassOptionsToPackage{hyphens}{url}
\PassOptionsToPackage{usenames, dvipsnames}{color}

\usepackage{orcidlink}
\renewcommand{\orcidID}[1]{\orcidlink{#1}}

\usepackage[linesnumbered,ruled,vlined]{algorithm2e}
\usepackage[hyphens]{url}
\usepackage{hyperref}
\usepackage{graphicx}
\usepackage{caption}
\usepackage{amsmath,amsfonts,bm}
\usepackage[utf8]{inputenc} %

\usepackage[usenames, dvipsnames]{color} %
\usepackage{xcolor} %

\definecolor{my-full-blue}{HTML}{1F77B4}

\definecolor{my-full-orange}{HTML}{FF7F0E}

\definecolor{my-full-green}{HTML}{2CA02C}

\definecolor{my-full-red}{HTML}{D62728}

\definecolor{my-full-purple}{HTML}{9467BD}

\definecolor{my-full-brown}{HTML}{8C564B}

\definecolor{my-full-olive}{HTML}{BCBD22}

\definecolor{my-full-cyan}{HTML}{17BECF}

\definecolor{my-full-gray}{HTML}{F7F7F7}

\colorlet{my-blue}{my-full-blue!30}
\colorlet{my-orange}{my-full-orange!30}
\colorlet{my-green}{my-full-green!30}
\colorlet{my-red}{my-full-red!30}
\colorlet{my-purple}{my-full-purple!30}
\colorlet{my-brown}{my-full-brown!30}
\colorlet{my-cyan}{my-full-cyan!30}
\colorlet{my-olive}{my-full-olive!30}

\usepackage{tikz}

\usetikzlibrary{arrows}
\usetikzlibrary{automata}
\usetikzlibrary{calc}
\usetikzlibrary{backgrounds}
\usetikzlibrary{decorations.markings}
\usetikzlibrary{decorations.pathmorphing}
\usetikzlibrary{decorations.pathreplacing}
\usetikzlibrary{fit}
\usetikzlibrary{patterns}
\usetikzlibrary{positioning}
\usetikzlibrary{shadows}
\usetikzlibrary{shapes}
\usetikzlibrary{shapes.misc}
\usetikzlibrary{math}
\usetikzlibrary{shapes.geometric}

\usepackage{subcaption}
\usepackage{adjustbox}
\usepackage{nth}
\usepackage{xparse}
\usepackage{multirow}
\usepackage{booktabs}
\usepackage{calc}
\usepackage{wrapfig}
\usepackage{pifont}
\usepackage{caption}
\usepackage{float}

\def\spec{\mathcal{I}}
\def\post{\psi}

\def\E{\mathbb{E}}
\def\va{{\bm{a}}}
\def\vb{{\bm{b}}}
\def\vc{{\bm{c}}}
\def\vd{{\bm{d}}}
\def\ve{{\bm{e}}}
\def\vh{{\bm{h}}}
\def\vn{{\bm{n}}}
\def\vx{{\bm{x}}}
\def\vz{{\bm{z}}}
\def\mA{{\bm{A}}}
\def\mC{{\bm{C}}}
\def\mD{{\bm{D}}}

\def\eps{{\epsilon}}
\DeclareMathOperator*{\argmax}{arg\,max}

\DeclareMathOperator{\diag}{diag}

\DeclareMathOperator{\supp}{supp}

\newcommand{\fcol}{0.65}
\newcommand{\colsep}{0.2}
\newcommand{\cmark}{\ding{51}\xspace}%
\newcommand{\xmark}{\ding{55}\xspace}%

\usepackage[capitalize]{cleveref}
\crefformat{section}{\S#2#1#3}
\crefrangeformat{section}{\S#3#1#4\crefrangeconjunction\S#5#2#6}
\crefmultiformat{section}{\S#2#1#3}{\crefpairconjunction\S#2#1#3}{\crefmiddleconjunction\S#2#1#3}{\creflastconjunction\S#2#1#3}
\newcommand{\crefrangeconjunction}{--}
\crefname{problem}{Problem}{Problems}
\crefname{algocf}{Algorithm}{Algorithms}
\crefname{line}{Line}{Lines}
\crefalias{algocfline}{line}
\newcommand{\sfref}[1]{(\subref{#1})}

\usepackage[firstpage]{draftwatermark}  %
\SetWatermarkAngle{0}
\SetWatermarkText{\raisebox{17.75cm}{%
  \hspace{0.1cm}%
  \href{https://doi.org/10.5281/zenodo.6551368}{\includegraphics{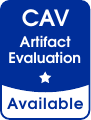}}%
  \hspace{9.1cm}%
  \includegraphics{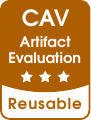}%
}}

\title{Shared Certificates for\\Neural Network Verification\protect\footnote[4]{Extended version of our CAV'22 paper. First published as Fischer et al. \cite{fischer2022shared}.}}
\author{
Marc Fischer\inst{1}\fnmsep\protect\footnote[1]{equal contribution}\orcidID{0000-0002-4157-1235}
\and
Christian Sprecher\inst{2}\fnmsep\protect\footnotemark[1]\fnmsep\protect\footnote[5]{work performed while at ETH Zurich}
\and\\
{Dimitar Iliev} Dimitrov\inst{1}\orcidID{0000-0001-9813-0900}
\and
Gagandeep Singh\inst{3}\orcidID{0000-0002-9299-2961}
\and
Martin Vechev\inst{1}\orcidID{0000-0002-0054-9568}
}
\institute{
ETH Zurich, Switerland,
\email{marc.fischer@inf.ethz.ch}, \email{dimitar.iliev.dimitrov@inf.ethz.ch}, \email{martin.vechev@inf.ethz.ch}
\and
Nostic Solutions AG, Switzerland, \email{christian.sprecher@nostic.ch}
\and
University of Illinois at Urbana-Champaign \& VMware Research, USA, \email{ggnds@illinois.edu}
}

\begin{document}

\maketitle

\begin{abstract}
Existing neural network verifiers compute a proof that each input is handled correctly under a given perturbation by propagating a symbolic abstraction of reachable values at each layer. This process is repeated from scratch independently for each input (e.g., image) and perturbation (e.g., rotation), leading to an expensive overall proof effort when handling an entire dataset. In this work, we introduce a new method for reducing this verification cost without losing precision based on a key insight that abstractions obtained at intermediate layers for different inputs and perturbations can overlap or contain each other. Leveraging our insight, we introduce the general concept of shared certificates, enabling proof effort reuse across multiple inputs to reduce overall verification costs.
We perform an extensive experimental evaluation to demonstrate the effectiveness of shared certificates in reducing the verification cost on a range of datasets and attack specifications on image classifiers including the popular patch and geometric perturbations.
We release our implementation at \url{https://github.com/eth-sri/proof-sharing}.

\keywords{Neural Network Verification
\and
Local Verification
\and
Adversarial Robustness
}
\end{abstract}

\section{Introduction}\label{sec:introduction} %

\begin{figure*}[t]
  \centering

  \newlength{\myheight}
  \newlength{\firstcol}
  \newlength{\secondcol}
  \setlength{\firstcol}{\textwidth*\real{\fcol}-\textwidth*\real{\colsep}*\real{0.5}}
  \setlength{\secondcol}{\textwidth-\firstcol-\textwidth*\real{\colsep}*\real{0.5}}

    \centering
    \adjincludegraphics[width=0.8\textwidth, gstore height=\myheight]{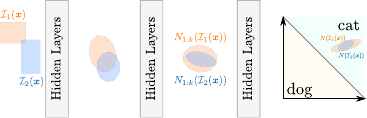}
    \label{fig:intro:verification}
   \caption{Visualization of neural network verification. The input regions $\spec_{1}(\vx), \spec_{2}(\vx)$ are propagated layer by layer through a neural network $N$. The high-dimensional convex shapes are visualized in 2d. While initially $\spec_{1}(\vx)$ and $\spec_{2}(\vx)$ only slightly overlap, at layer $k$, $N_{1:k}(\spec_{2}(\vx))$ is fully contained in $N_{1:k}(\spec_{1}(\vx))$.
}
    \label{fig:intro}
\end{figure*}

The success of neural networks across a wide range of application domains \cite{KrizhevskySH17,AlphaGo} has led to their widespread application and study. Despite this success, neural networks remain vulnerable to adversarial attacks~\cite{Brown17Patches,MadryTraining} which raises concerns over their trustworthiness in safety-critical settings such as autonomous driving and medical devices. To overcome this barrier, formal verification of neural networks has been proposed as a key technology in the literature~\cite{trustworthyacm}.
As a result, recent years have witnessed a growing interest in verifying critical safety properties of neural networks (e.g., fairness,  robustness) \cite{AI2,Reluplex,Marabou:19,SinghGMPV18,DeepPoly,WongK18,Zhang2018} specified using pre and post conditions over network inputs and outputs respectively. Conceptually, existing verifiers propagate sets of inputs in the precondition captured in symbolic form (e.g., convex sets) through the network, an expensive process that produces over-approximations of all possible values at intermediate layers. The final abstraction of the output can then be used to check postconditions.
The key technical challenge all existing verifiers aim to address is speeding up and scaling the certification process, i.e, faster and more efficient propagation of symbolic shapes while reducing the overapproximation error.%
\paragraph{This work: accelerating certification via proof sharing.}
In this work, we propose a new, complementary method for accelerating neural network verification based on the key observation that instead of treating each certification attempt in isolation as existing verifiers do, we can reuse proof effort among multiple such attempts, thus obtaining significant overall speed-ups without losing precision. \cref{fig:intro} illustrates both, standard verification and the concept of proof sharing.

In standard verification an input region  $\spec_1(\vx)$ (orange square) is propagated from left to right, obtaining intermediate shapes at each intermediate layer (here the goal is to verify all points in the input region are classified as ``cat'' by the neural network $N$). 
We observe that the abstraction obtained for a new region $\spec_2(\vx)$ (e.g., blue shapes) can be contained inside existing abstractions from $\spec_1(\vx)$, an effect we term \emph{proof subsumption}.
This effect can be observed both between abstractions obtained from different specifications (e.g., $\ell_{\infty}$ and adversarial patches) for the same data point and between proofs for the same property but different, yet semantically similar inputs.
Building on this observation, we introduce the notion of proof sharing via templates. Proof sharing works in two steps: first, we leverage abstractions from existing proofs in order to create templates, and second, we augment the verifier with these templates, stopping the expensive propagation at an intermediate layer as soon as the newly generated abstraction is included inside an existing template. Key technical ingredients to the effectiveness of our approach are fast template generation and inclusion checking techniques. We experimentally demonstrate that proof sharing can achieve significant speed-ups in challenging scenarios including proving robustness to adversarial patches \cite{ChiangNAZSG20} and geometric perturbations \cite{DeepG} across different neural network architectures.

\paragraph{Main Contributions} %
Our key contributions are:
\begin{itemize}
	\item An introduction and formalization of the concept of proof sharing in neural network verification: the idea that some proofs capture others (\cref{sec:transfer}).
	\item A general framework leveraging the above concept, enabling proof effort reuse via proof templates (\cref{sec:template-generation-individual}).
	\item A thorough experimental evaluation involving verification of neural network robustness against challenging adversarial patch and geometric perturbations, demonstrating that our methods can achieve proof match rates of up $95 \%$ as well as provide non-trivial end-to-end certification speed-ups (\cref{sec:evaluation}). 
\end{itemize}

\section{Background} \label{sec:background}

Here we formally introduce the necessary background for proof sharing.

\paragraph{Neural Network}
A neural network $N$ is a function $N: \mathbb{R}^{d_\text{in}} \to \mathbb{R}^{d_\text{out}}$, commonly built from individual layers $N = N_L \circ N_{L-1} \circ \cdots \circ N_1$. 
Throughout this text, we consider feed-forward neural networks, where each layer $N_i(\vx) = \max(\mA \vx + \vb, 0)$ consists of an affine transformation ($\mA \vx + \vb$) as well as a rectified linear unit (ReLU), that applies the $\max$ with 0 elementwise. A neural network, classifying inputs into $c$ classes, outputs $d_\text{out} := c$ scores, one for each class, and assigns the class with the highest score as the predicted one.
While, as is common in the neural network verification literature, we use image classification as a proxy task, many other applications work analogously.
Our approach also naturally extends to other types of neural networks, if verifiers exist for these architectures. We discuss the challenges and limitations of such generalizations in \cref{sec:requirements-for-proof-sharing}.
In the following, for $k < L$, we let $N_{1:k}$ denote the application of the first $k$ layers and $N_{k+1:L}$ denote the last $L-k$ layers respectively.

\paragraph{(Local) Neural Network Verification}
Given a set of inputs and a postcondition $\post$, the goal of neural network verification is to prove that $\post$ holds over the output of the neural network corresponding to the given set of inputs. In this work, we focus on local verification, proving that $\post$ holds for the network output for a given region $\spec(\vx) \subseteq \mathbb{R}^{d_\text{in}}$ formed around the input $\vx$. Formally, we state this as:
\begin{problem}[Local neural network verification] \label{prob:verification}
For a region $\spec(\vx) \subseteq \mathbb{R}^{d_\text{in}}$, neural network $N$, and postcondition $\post$, verify that
$\forall \vz \in \spec(\vx). \; N(\vz) \models \post$. We write $\spec(\vx) \models \post$ if  $\forall \vz \in \spec(\vx). \; N(\vz) \models \post$.
\end{problem}

Here, we restrict ourselves to verifiers based on abstract interpretation \cite{Cousot:77,AI2} as they achieve state-of-the-art precision and scalability~\cite{DeepPoly,SinghGMPV18}. Further, many other popular verifiers~\cite{Weng2018,Zhang2018} can be formulated using abstract interpretation. These verifiers propagate $\spec(\vx)$ symbolically through the network $N$ layer-by-layer using abstract transformers, which overapproximate the effect of applying the transformations defined in the different layers on symbolic shapes. The propagation yields an abstraction of the exact shape at each layer. The verifiers finally check if the abstracted output implies $\post$. 
This is showcased in \cref{fig:intro}, where the input regions $\spec_{1}(\vx)$ and $\spec_{2}(\vx)$ are propagated layer-by-layer through $N$.

For a verifier $V$, we let $V(\spec(\vx), N)$ denote the abstraction obtained after the propagation of $\spec(\vx)$ through the network $N$. We declutter notation by overloading $N$ and writing $N(\spec(\vx))$ for the same if $V$ is clear from context, i.e., $V(\spec(\vx), N) = N(\spec(\vx))$.

We consider robustness verification, where the goal is to prove that the network classification does not change within an input region. A common input region is the $\ell_{\infty}$-bounded additive noise, defined as $\spec_{\epsilon}(\vx) := \{\vz \mid \| \vx - \vz \|_{\infty} \leq \epsilon \}$. Here, $\epsilon$ defines the size of the maximal perturbation to $\vx$. The postcondition $\post$ denotes classification to the same class as $\vx$. Throughout this paper, we consider different instantiations for $\spec(\vx)$ but assume that $\post$ denotes classification invariance (although other choices would work analogously). Due to this, we refer to $\spec(\vx)$ as input region and specification interchangeably.
For example, in \cref{fig:intro}, the goal is to verify that all points contained in $N(\spec_{1}(\vx))$ are classified as ``cat''.

\section{Proof Sharing with Templates}\label{sec:transfer} %
Before introducing our framework for proof sharing, we further expand the motivation example discussed in \cref{fig:intro}.

\subsection{Motivation: Proof Subsumption} \label{sec:transfer:motivation}
As stated earlier, we empirically observed that for many input regions $\spec_i(\vx)$ and $\spec_j(\vx)$, the abstraction corresponding to one region at some intermediate layer $k$ contains that of another. Formally: 
\begin{definition}[Proof Subsumption]
For specifications $\spec_i(\vx), \spec_j(\vx)$, we say that the proof of $\spec_i(\vx)$ subsumes that of $\spec_j(\vx)$ if at some layer $k$, $N_{1:k}(\spec_j(\vx)) \subseteq N_{1:k}(\spec_i(\vx))$, which we denote as $\spec_j(\vx) \subseteq_{N,k} \spec_i(\vx)$.
\end{definition}

\input{figures/patch_example}
While not formally required, particularly interesting are cases where proof subsumption occurs despite $\spec_i(\vx) \not \subseteq \spec_j(\vx)$. This form of proof subsumption is showcased in \cref{fig:intro}, where $\spec_{1}(\vx)$ and $\spec_{2}(\vx)$ have only a small overlap, yet $\spec_2(\vx) \subseteq_{N,k} \spec_1(\vx)$.
For another example, consider a neural network $N$ trained as a hand-written digit classifier for the MNIST dataset \cite{MNIST} (example shown in \cref{fig:patch}) and the following two specifications:
\begin{itemize}	
	\item $\ell_\infty$-bounded perturbations: all the pixels in an input image can arbitrarily be changed independently by a small amount $\spec_{\epsilon}(\vx) := \{\vz \mid \| \vx - \vz \|_{\infty} \leq \epsilon \}$, 
\end{itemize}	

\begin{itemize}	
	\item adversarial patches \cite{ChiangNAZSG20}. A $p \times p$ patch inside which the pixel intensity can vary arbitrarily is placed on an image at coordinates $(i,j)$, for which we write $\spec_{p \times p}^{i,j}$. We showcase a patch in \cref{fig:patch} and formally define them in \cref{sec:robustn-advers-patch}. %
\end{itemize}

\begin{figure}[t]
\begin{minipage}{\textwidth}
	\begin{minipage}[b]{0.6\textwidth}
		\centering
		\captionof{table}{Proof subsumption on a robust MNIST classifier with 94 \% accuracy. Verif. acc. denotes the percentage of verifiable inputs from the test set for $\ell_{\infty}$-perturbations ($\spec_{\eps}$).}
		\label{tab:motivation}
		\vspace{2mm}
		\begin{tabular}[b]{@{}c@{\hskip 0.3cm}c@{\hskip 0.3cm}cccccc@{}} %
		\toprule
			\multirow{2}{*}{$\eps$} & \multirow{2}{*}{\shortstack[c]{verif. acc.\\for $\spec_{\eps}$ [\%]}} &  \multicolumn{5}{c}{ $\spec_{2\times 2}^{i,j}(\vx) \subseteq_{N,k} \spec_{\eps}(\vx)$ at layer k [\%]}\\
		\cmidrule(lr){3-7}
		   &   & 1 & 2 & 3 & 4 & 5\\
		\midrule
		$0.1$ & 89.74& 61.40 & 72.85 & 77.65 & 81.75 & 82.70\\
		$0.2$ & 81.40& 62.85 & 77.05 & 82.40 & 86.05 & 86.60\\
		\bottomrule
		\end{tabular}
	\end{minipage}
	\hfill
	\begin{minipage}[b]{0.35\textwidth}
	\begin{center}
		\includegraphics[width=0.8\textwidth]{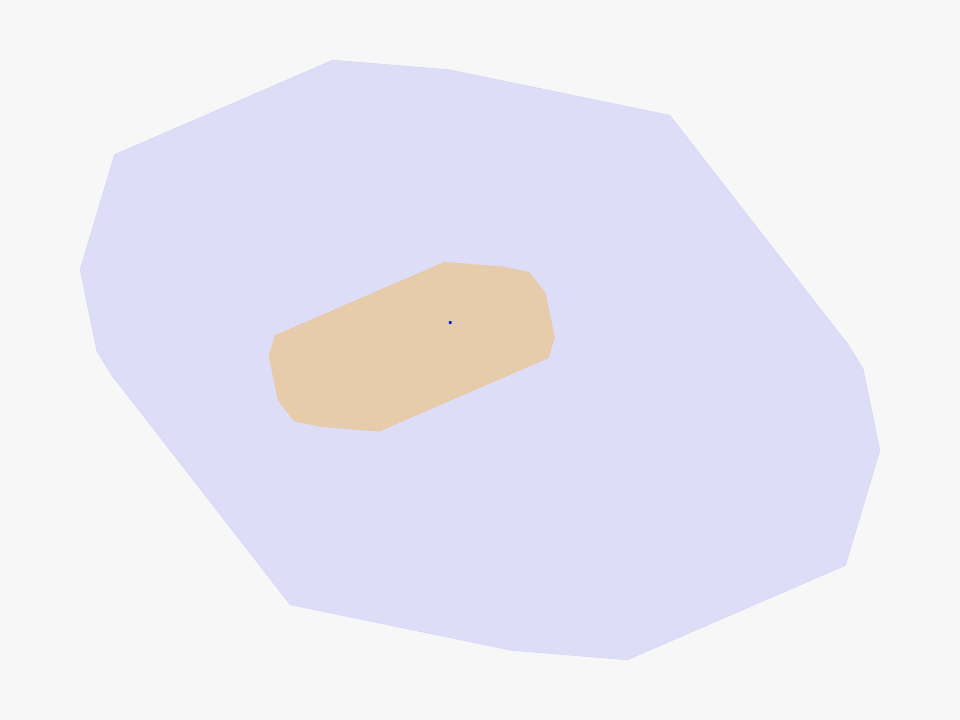}
		\captionof{figure}{The abstraction obtained for
			$\spec_{\epsilon}(\vx)$ (blue) contains that for
			$\spec^{i,j}_{2 \times 2}(\vx)$ (orange) (projected to $d=2$).}
		\label{fig:motivation:subsumption}
	\end{center}
	\end{minipage}
\end{minipage}
\end{figure}

Clearly $\spec_{p \times p}^{i,j}(\vx) \not\subseteq \spec_{\epsilon}(\vx)$ (unless $\eps=1$).
In \cref{tab:motivation},  we show that for a classifier (5 layers with 100 neurons each) we indeed observe proof subsumption. We report the accuracy, i.e., the rate of correct predictions on the unperturbed test data, as well as the certified accuracy, i.e., the rate of samples $\vx$ for which the prediction is correct and $\spec(\vx) \models \post$ is verified, for $\spec_{\eps}$ with $\eps=0.1$ and $0.2$ over the whole test set.
We also show the percentage of $\spec_{2 \times 2}^{i,j}(\vx)$ contained in $\spec_{\eps}(\vx)$ at layer $k$. To this end, we pick 1000 random $\vx$ for which $\spec_{\eps}(\vx)$ is verifiable and sample 2 $(i, j)$ pairs each. We utilize a Box domain verifier and a robustly trained network \cite{DiffAI}.
\cref{fig:motivation:subsumption} shows a patch specification $\spec_{2 \times 2}^{i,j}(\vx)$ (in orange) contained in the $\ell_{\infty}$ specification $\spec_\epsilon$ (in blue) projected to 2 dimensions via PCA.

\paragraph{Reasons for Proof Subsumption}
In \cref{tab:motivation}, we observe that the rate of proof subsumption increases with larger $\eps$ and $k$. These observations give an intuition as to why we observe proof subsumption.
First, as input regions pass through the neural network, in each layer the abstractions become more imprecise. While this fundamentally limits verification, it makes the subsumption of abstractions more probable. This effect increases, when increasing $\eps$ for $\spec_\eps$.
Second, and more fundamentally, while passing through the layers of a neural network, we observed that semantically similar yet distinct image inputs, e.g., two similar-looking handwritten digits, have activation vectors that grow closer in $\ell_{2}$ norm as they pass through the layers of the neural network \cite{KrizhevskySH17,SzegedyIntriguing}. This effect is a consequence of the neural network distilling low-level information (e.g., individual pixel values) into high-level concepts (e.g., the classes of digits). As specifications (and their proofs) correspond to sets of concrete inputs, a similar effect may apply.
We conjecture that these two effects drive the observed proof subsumption.

\begin{figure*}[ht]
  \centering

  \setlength{\firstcol}{\textwidth*\real{\fcol}-\textwidth*\real{\colsep}*\real{0.5}}
  \setlength{\secondcol}{\textwidth-\firstcol-\textwidth*\real{\colsep}*\real{0.5}}

  \begin{subfigure}{\firstcol}
    \centering
    \adjincludegraphics[width=\textwidth, gstore height=\myheight]{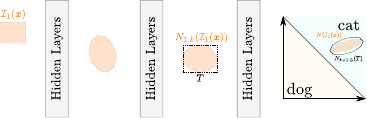}
    \subcaption{We generate a (verifiable) template $T$ from the abstraction obtained by propagating the orange input.}
     \label{fig:template:template}
  \end{subfigure}
  \hfill
  \begin{subfigure}{\secondcol}
    \centering
    \includegraphics[height=\myheight]{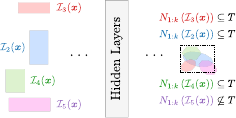}
    \subcaption{We shortcut the verification if intermediate abstraction are contained in the $T$.}
     \label{fig:template:template_matching}
  \end{subfigure}

  \caption{
    Conceptualization of
    proof sharing with templates.
    In \sfref{fig:template:template}
    we create a verifiable template $T$ (black-dashed border) from specification $N_{1:k}(\spec_{1}(\vx))$.
	When verifying new specifications $\spec_{2}, \dots, \spec_{5}$, shown in \sfref{fig:template:template_matching}, we can shortcut the verification of all but $\spec_5$ by subsuming them in $T$.
 }
    \label{fig:template}
\end{figure*}

\subsection{Proof Sharing with Templates} \label{sec:transfer:transfer}
Leveraging this insight, we introduce the idea of proof sharing via templates, showcased in \cref{fig:template}.
We use an abstraction obtained from a robustness proof $N_{1:k}(\spec_{1}(\vx))$ at layer $k$ to create a template $T$. After ensuring that $T$ is verifiable, it can be used to shortcut the verification of other regions, e.g., of $\spec_{2}(\vx), \dots, \spec_{5}(\vx)$.
Formally we decompose proof sharing into two sub-problems: (i) the generation of proof templates and (ii) the matching of abstractions corresponding to other properties to these templates.
For simplicity, here we only consider templates at a single layer $k$ of the neural network and we show an extension to multiple layers in \cref{sec:robustn-advers-patch}.

Our goal is to construct a template $T$ at layer $k$ that implies the postcondition and captures abstractions at layer $k$ obtained from propagating several $\spec_i(\vx)$. As it is challenging to find a single $T$ that captures abstractions corresponding to many input regions, yet remains verifiable, we allow a set of templates $\mathcal{T}$. We state this formally as:
\begin{problem}[Template Generation] \label{prob:generation}
	For a given neural network $N$, input $\vx$ and set of specifications $\spec_1, \dots, \spec_r$, layer $k$ and a postcondition $\post$,
  find a set of templates $\mathcal{T}$ with $|\mathcal{T}| \leq m$ such that:
\begin{align}
	&\argmax_{\mathcal{T}} \sum_{i=1}^r  \left[\bigvee\limits_{T \in \mathcal{T}} N_{1:k}(\spec_i(\vx)) \subseteq T  \right] \label{eq:generation-opt} \\
&\text{s.t.}\;\forall \; T \in \mathcal{T}. \, N_{k+1:L}(T) \models \post \ .\nonumber
\end{align}
\end{problem}
Intuitively, \cref{eq:generation-opt} aims to find a set $\mathcal{T}$ of templates $T$ at layer $k$, such that the maximal amount (via the sum) of specifications $\spec_1, \dots, \spec_r$ is contained in at least one template $T$ (via the disjunction) while ensuring that the individual $T$ are still verifiable (via the constraint on the second line).
As neural network verification required by the constraints of \cref{eq:generation-opt}, is NP-complete \cite{Reluplex}, computing an exact solution to \cref{prob:generation} is computationally infeasible.
Therefore, we compute an approximate solution to \cref{eq:generation-opt}. In general, \cref{prob:generation} does not necessarily require that the templates $T$ are created from previous proofs. However, building on proof subsumption, as discussed in \cref{sec:transfer:motivation}, in \cref{sec:template-generation-individual} we will infer the templates from previously obtained abstractions.

To leverage proof sharing once the templates $\mathcal{T}$ are obtained, we need to be able to match an abstraction $S = N_{1:k}(\spec(\vx))$ verified using proof transfer to a template in $\mathcal{T}$:

\begin{problem}[Template Matching] \label{prob:matching} Given a set
  of templates $\mathcal{T}$ at layer $k$ of a neural network $N$,
  and a new input region $\spec(\vx)$, determine whether
  there exists a $T \in \mathcal{T}$ such that
  $S \subseteq T$, where $S = N_{1:k}(\spec(\vx))$.
\end{problem}

Together, \cref{prob:generation,prob:matching} outline a general
framework for proof sharing, permitting many instantiations.
We note that \cref{prob:generation,prob:matching} present an inherent precision vs. speed trade-off: \cref{prob:matching} can be solved most efficiently for small values of $m = |\mathcal{T}|$ and simpler representations of $T$ (allowing faster checking of $S \subseteq T$) at the cost of lower proof matching rates. Alternatively, \cref{eq:generation-opt} can be maximized by large $m$ and $T$ represented by complex abstractions, thus attaining high precision but expensive template generation and matching.

\paragraph{Beyond proof sharing on the same input}
In this section, we focused on proof sharing for different specifications of the same input $\vx$. However, we observed that proof sharing is even possible between specifications defined on different inputs $\vx$ and $\vx'$. To facilitate the use of templates in this setting, \cref{eq:generation-opt} in \cref{prob:generation} can be adapted to consider an input distribution.
We provide an investigation along these lines in \cref{sec:offline}.

\section{Efficient Verification via Proof Sharing}\label{sec:template-generation-individual} %
We now consider an instantiation of proof sharing
where we are given an input $\vx$ and properties $\spec_{1}, \dots ,\spec_{r}$ to verify.
Our general approach, based on \cref{prob:generation,prob:matching}, is shown in \cref{alg:verif_online}. In this section, we first discuss \cref{alg:verif_online} in general. We then describe the possible choices of abstract domains and their implications on the algorithm, followed by a discussion on template generation for two different specific problems. Finally, we conclude the section with a discussion on the conditions for effective proof sharing verification. 

In \cref{alg:verif_online},  we first create the set of templates $\mathcal{T}$ (\cref{alg:verif_online:gen}, discussed shortly) and subsequently verify $\spec_{1}, \dots ,\spec_{r}$ using $\mathcal{T}$.
Here, we consider two, potentially identical, verifiers $V_T$ and $V_S$, where $V_T$ is used to create the templates $\mathcal{T}$ and $V_S$ is used to propagate input regions up to the template layer $k$. For each $\spec_{i}$ we propagate it up to layer $k$ (\cref{alg:verif_online:match_start}) to obtain $S = N_{1:k}(\spec_{i}(\vx))$ and check if we can match it to a template $T_j \in \mathcal{T}$ (\cref{alg:verif_online:incl}) using an inclusion check. If a match is found, then we conclude that $N(\spec_{i}(\vx)) \models \post$ and set the verification output $v_i$ to True. If this is not the case (\cref{alg:verif_online:check}) we verify $N(\spec_{i}(\vx)) \models \post$  directly by checking $V_S(S, N_{k+1:L}) \models \post$.
If the template generation fails, we revert to verifying $\spec_i$ by applying $V_S$ in the usual way (omitted in \cref{alg:verif_online}).

\paragraph{Soundness}
As long as the templates $T$ are sound, this procedure is sound, i.e \cref{alg:verif_online} only returns $v_i = \text{True}$ if $\forall \vz \in \spec_{i}(\vx). \; N(z) \models \post$ holds. Formally:

\begin{theorem} \label{thm:soundness}
	\cref{alg:verif_online} is sound if $\forall \; T \in \mathcal{T}\!,\, z \in T.\;  N_{k+1:L}(z) \models \post$ and $V_S$ is sound.
\end{theorem}

\begin{wrapfigure}[21]{R}{0.61\textwidth}
\vspace{-1mm}
\centering
\begin{minipage}{0.6\textwidth}
\begin{algorithm}[H]
\SetInd{0.425em}{0.4em}
\DontPrintSemicolon
\SetAlgoLined
\SetNoFillComment
\SetKwInOut{KwInput}{Input}
\SetKw{KwBreak}{break}
\KwInput{$\vx, \spec_{1}, \dots, \spec_{r}, k, \psi$, verifiers $\!V_S, V_T$}
\KwResult{$v_{1}, \dots, v_{r}$ indicating $v_{i} := \left( N(\spec_{i}(\vx)) \models \psi \right)$}

	$\mathcal{T} \!\gets\!$ \textsc{gen\_templates}$(\vx, N, k, \psi, V_S, V_T)$\; \nllabel{alg:verif_online:gen}
$v_{1}, \dots, v_{r} \gets $ False\;
\For{$i \gets 1$ \KwTo $r$}{
  $S \gets V_S(\spec_{i}(\vx), N_{1:k})$\;  \nllabel{alg:verif_online:match_start}
  \For{$T_{j} \in \mathcal{T}$ }{ \nllabel{alg:verif_online:tincl_start}
    \If{$S \subseteq T_{j}$}{  \nllabel{alg:verif_online:incl}
      $v_{i} \gets $ True \;
      \KwBreak\;
    }
  }   \nllabel{alg:verif_online:match_end}
  \If{$\lnot v_{i}$}{ \nllabel{alg:verif_online:check}
	$v_{i} \gets \left( V_S(S, N_{k+1:L}) \models \psi \right)$ \; \nllabel{alg:verif_online:post}
  }
}
\Return{$v_{1}, \dots, v_{r}$} \nllabel{alg:verif_online:return}
\caption{Neural Network Verification Utilizing Proof Templates}
\label{alg:verif_online}
\end{algorithm}
\end{minipage}
\end{wrapfigure}

This holds by the construction of the algorithm: 

\begin{proof}
	For a given $\vx$ and $\spec_i$, \cref{alg:verif_online} only claims $v_i = \text{True}$ if either the check in (i) \cref{alg:verif_online:incl} or (ii) \cref{alg:verif_online:check} succeeds.
	Since $V_S$ is sound, we know that $\forall \vz \in \spec_{i}(\vx). \; N_{1:k}(z) \in S$. 
	Therefore in case (i) by our requirement on $T$ as well as $S \subseteq T$ it follows that  $\forall \vz \in \spec_{i}(\vx). \; N(z) \models \post$.
	In case (ii) we execute \cref{alg:verif_online:post} and the same property holds due to the soundness of $V_S$.
\end{proof}

Importantly, \cref{thm:soundness} shows that the generation process of $\mathcal{T}$  does not affect the overall soundness as long as the set of templates $\mathcal{T}$ fulfills the condition in \cref{thm:soundness}. 
In particular, that means that when solving \cref{prob:generation}, it suffices to show the side condition $(\forall \; T \in \mathcal{T}. \, N_{k+1:L}(T) \models \post)$ holds, while heuristically approximating the actual optimization criteria.
We let $V_T$ denote the verifier used to ensure this property in \textsc{gen\_templates}.

\paragraph{Precision}
We say a verifier $V_1$ is more precise than another verifier $V_2$ on $N$ 
if out of a set of specifications it can verify some that $V_2$ can not.

\begin{theorem} \label{thm:precission}
	If $V_S(V_S(\spec_i(\vx), N_{1:k}), N_{k+1:L}) = V_S(\spec_i(\vx), N)$, then \cref{alg:verif_online} is at least as precise as $V_S$. 
\end{theorem}

\begin{proof}
	Since, even if the inclusion check in \cref{alg:verif_online:incl} fails, 
	due to \cref{alg:verif_online:post} we output $v_i = V_S(V_S(\spec_i(\vx), N_{1:k}), N_{k+1:L}) \models \post$ (\cref{alg:verif_online:post}), which by our requirement equals $v_i = V_S(\spec_i(\vx), N) \models \post$. Therefore we have at least the precision of $V_S$.
\end{proof}

The required property holds for any verifier $V_S$ for which the abstractions of all network layers depends only on the abstractions from previous layers and is fulfilled for all verifiers considered in this paper. For verifiers $V_S$ that do not fulfill the required property, potential losses in precision can be remedied (at the cost of runtime) by using $V_S(\spec_i(\vx), N_{1:L})$ in \cref{alg:verif_online:post}.
Interestingly, it is even possible to increase the precision of \cref{alg:verif_online} over $V_S$ by creating templates $T$ that are verified with a more precise verifier $V_T$. However, in this discussion, we restrict ourselves to speed gains. We believe that obtaining precision gains requires instantiating our framework with a significantly different approach than that taken for improving speed which is the main focus of our work. We leave this as an interesting item for future work.

\paragraph{Run-Time}
Here, we aim to characterize the run-time of \cref{alg:verif_online} as well as its speed-up over conventional verification.
For an input $\vx$, (keeping the other parameters fixed), the expected run time is
\begin{equation}
	\label{eq:runtime}
	t_{PS} = t_{\mathcal{T}} + r (t_{S} + t_{\subseteq} + (1-\rho) t_{\post})
\end{equation}
where $t_{\mathcal{T}}$ is the expected time required to generate the templates at \cref{alg:verif_online:gen},
$r$ is the number of specifications to be verified, $t_{S}$ is the expected time to compute $S$ (\cref{alg:verif_online:match_start}), $t_{\subseteq}$ is the time to check $S \subseteq T$ for $T \in \mathcal{T}$ until a match is found (\cref{alg:verif_online:tincl_start}~to~\cref{alg:verif_online:match_end}), $\rho \in [0, 1]$ is the rate of specifications where a template is found and $t_{\post}$ is the time required to check $\post$ on the network output corresponding to $S$ (\cref{alg:verif_online:post}). 
This time is minimized if the individual expected run times $t_{\mathcal{T}}, t_S, t_{\post}$ are minimal and $\rho$ is large (i.e., close to 1).
Unfortunately, computing the template match rate $\rho$ analytically is challenging and requires global reasoning over the neural network for all valid inputs, which are not clearly defined. However, our empirical analysis (in \cref{sec:evaluation}) shows that $\rho$ is higher when templates are created at later layers (as in \cref{sec:transfer:motivation}).

To determine the speed-up compared to a baseline standard verifier, we make the simplifying assumption that there is a single verifier $V = V_S = V_T$ that has expected run-time $\nu$ for each layer. Thus, the expected run-time for the conventional verifier is $t_{BL} = rL\nu$. 
We have $t_{\mathcal{T}} = \lambda mL\nu$, $t_S = k\nu$, $t_{\post} = (L - k)\nu$, $t_{\subseteq} = \eta m$ and ultimately $t_{PS} = (m+r(1-\rho))L\nu + r \rho k\nu + r \eta m$ for constants $\lambda \in \mathbb{R}_{>0}$, which indicates the overhead in generating one template over just verifying it, and $\eta \in \mathbb{R}_{>0}$ which denotes the time required to perform an inclusion check for one template.
As this phrasing shows, \cref{alg:verif_online} has the same asymptotic runtime as the base verifier $V$.
Further, this formulation allows us to write our expected speed-up as $\tfrac{t_{BL}}{t_{PS}} = \tfrac{r}{\lambda m + \eta rm/L\mu + r \rho k/L + r(1-\rho)}$. 
This speed-up is maximized when $k$ is small compared to $L$, i.e., templates are placed early in the neural network, the matching rate $\rho$ is close to 1, and $m, \lambda, \eta$ are small, i.e., generation and matching are fast.
Unfortunately, these requirements are at odds with each other: as we show in \cref{sec:evaluation}, higher $m$ leads to higher matching rate $\rho$ and $\rho$ is naturally higher for templates later in the neural network (higher $k$).
Thus high speed-ups require careful hyper-parameter choices.

To showcase how we can achieve good templates as well as fast matching, we next discuss the choice of the abstract domain to be used in the propagation and the representation of the templates. Then we discuss the template generation procedure and instantiate it for the verification of robustness to adversarial patches and geometric perturbations.

\subsection{Choice of Abstract Domain}\label{sec:domains}
To solve \cref{prob:generation,prob:matching} in a way that minimizes the expected runtime and maximizes the overall precision, the choice of abstract domain is crucial. 
Here we briefly review common choices of abstract domains for neural network verification and how they are suited to our problem.
Geometrically these domains can be thought of as a convex abstraction of the set of vectors representing reachable values at each layer of the neural network.
We say that an abstraction $a_1$ is more precise than another abstraction $a_2$, if and only if $a_1 \subseteq a_2$, i.e, all points in $a_1$ occur in $a_2$. Similarly, we say that a domain is more precise than another if it can express all abstractions in the other domain.

The Box (or Interval) domain \cite{AI2,DiffAI,IBP} abstracts sets in $d$ dimensions 
as $B = \{\va + \diag(\vd) \ve \mid \ve \in [-1, 1]^{d} \}$ with center $\va \in \mathbb{R}^{d}$ and width $\vd \in \mathbb{R}_{\geq 0}^{d}$.
The Zonotope domain \cite{GoubaultP15,AI2,SinghGMPV18,WongK18,DiffAI} uses relaxations $Z$ of the form 
\begin{equation}
  \label{eq:zonotope}
  Z = \{\va + \mA \ve \mid \ve \in [-1, 1]^{q} \},
\end{equation}
parametrized with $\va \in \mathbb{R}^{d}$ and
$\mA \in \mathbb{R}^{d \times q}$.

\begin{wrapfigure}[13]{r}{0.35\textwidth}
\begin{center}
\bgroup
\def\arraystretch{1.5}%
\setlength\tabcolsep{2mm}
\vspace{-1.7cm}
\begin{tabular}{@{}ll|c|c|c|c@{}}
	\multicolumn{2}{c}{} &  \multicolumn{4}{c}{$T$} \\
	&& $B$ & $Z$ & $\alpha(Z)$ & $P$ \\ \cline{2-6}
	\multirow{3}{*}{$S$} & $B$ & \cmark & \xmark & \cmark & (\cmark) \\ \cline{2-6}
	& $Z$ & \cmark & \xmark & \cmark & \xmark \\ \cline{2-6}
	&$P$ & \cmark & \xmark & \cmark & (\cmark)  \\
\end{tabular}
\egroup
\end{center}
\captionof{table}{Feasibility of $S \subseteq T$ for Box $B$, Zonotope $Z$ (with order reduction) and DP Polyhedra $P$.}
\label{tab:inclusion}
\end{wrapfigure}

A third common choice are (restricted) convex Polyhedra $P$ \cite{Cousot:78,DeepPoly,Zhang2018}. Here, we consider $P$ to be in the DeepPoly (DP) domain \cite{DeepPoly,Zhang2018}.
Generally, Boxes are less precise, i.e. certify fewer properties, than Zonotopes or Polyhedra.

For efficient proof sharing, we require a fast inclusion check $S \subseteq T$, which is challenging in our context due to the high dimensionality $d$ of the intermediate neural network layers.
While we point the interested reader to \cite{SadraddiniT19} for a detailed discussion, we summarize the key results in \cref{tab:inclusion}.  There, \cmark denotes feasibility, i.e. low polynomial runtime (usually $2d$ comparisons, sometimes with an additional matrix multiplication), \xmark denotes infeasibility, e.g. exponential run time. 
If $T$ is a Box all checks are simple as it suffices to compute the outer bounding box of $S$ and compare the $2d$ constraints. 
If $T$ is a DP Polyhedra these checks require a linear program (LP) to be solved.
While the size of this LP permits a low theoretical time complexity, in case $S$ is a Box or DP Polyhedra, in practice, we consider calling an LP solver too expensive (denoted as (\cmark)).
For Zonotopes these checks are generally infeasible, as they require enumeration of the faces or corners, which is computationally expensive for large $d$ and $P$.
While Zonotopes can be encoded as Polyhedra (but not necessarily DP Polyhedra) and the same LP inclusion check as for $P$ could be used, the resulting LP would require exponentially many variables due to the previously mentioned enumeration.
However, by placing constraints on the matrix $\mA$ in \cref{eq:zonotope} these inclusion checks can be performed efficiently.
The mapping of a Zonotope to such a restricted Zonotope is called order reduction via outer-approximation \cite{SadraddiniT19,KopetzkiSA17}.

In particular, for a Zonotope $Z$ we consider the order reduction $\alpha_{\text{Box}}$  to its outer bounding box  (where $\mA$ is diagonal) and note that other choices of $\alpha$ are possible (e.g. the reduction to affine transformations of a hyperbox).

For a general Zonotope $Z$ its outer bounding box $Z' = \alpha_{\text{Box}}(Z)$ can be easily obtained. The center of $Z'$ is $\va$, the center of Z. The width $\vd \in \mathbb{R}_{\geq 0}^{d}$ is given as $d_{i} = \sum_{j=1}^{q} |A_{i,j}|$.
$Z'$ is represented as either a Box or a Zonotope (with $\mA = \diag(\vd)$).
To check $S \subseteq Z'$ for a general Zontope $S$ it suffices to check $\alpha_{\text{Box}}(S) \subseteq Z'$ which reduces to the simple inclusion check for boxes.

Based on the above discussion we will use the Zonotope domain to represent all abstractions,
and use verifiers $V_S = V_T$ that propagate these zonotopes using the state-of-the-art DeepZ transformers 
\cite{SinghGMPV18}.
To permit efficient inclusion checks we apply $\alpha_\text{Box}$ on the resulting zonotopes
to obtain the Box templates $T$, which we treat as a special case of Zonotopes.

\subsection{Template Generation}\label{sec:template-generation-individual-generation} %

We now discuss instantiations for \textsc{gen\_templates} in \cref{alg:verif_online}.
Recall from \cref{sec:transfer:motivation} the idea of proof subsumption, i.e. that abstractions for some specification contain abstractions for other specifications. Building on this, we relax the \cref{prob:generation} in order to create $m$ templates $T_{j}$ from intermediate abstractions $N_{1:k}(\hat{\spec}_{i}(\vx))$ for some $\hat{\spec}_{1}, \dots, \hat{\spec}_{m}$. 
Note that $\hat{\spec}_{j}$ are not necessarily directly related to the specifications $\spec_{1}, \dots ,\spec_{r}$ that we want to verify.
For a chosen layer $k$, input $\vx$, number of templates $m$ and verifiers $V_S$ and $V_T$ we optimize

\begin{equation}
  \label{eq:individual-template}
\begin{aligned}
	&\argmax_{\hat{\spec}_{1}, \dots, \hat{\spec}_{m} } \sum_{i=1}^r \left[ \bigvee\limits_{j = 1}^{m} V_S(\spec_i(\vx),N_{1:k}) \subseteq T_{j}  \right]\\
	&\text{where}\; T_{j} = \alpha_{\text{Box}}(V_T(\hat{\spec}_{j}(\vx), N_{1:k}))\\
	&\text{s.t.}\; V_T(T_{j}, N_{k+1:L}) \models \psi \;\text{for}\; j \in {1, \dots, m}.
\end{aligned}
\end{equation}

As originally in \cref{prob:generation} (\cref{eq:generation-opt}) we aim to find a set of templates such that the intermediate shapes at layer $k$ for most of the $r$ specifications are covered by at least one template $T$. In contrast to \cref{eq:generation-opt}, we tie $T_j$ to the specifications $\hat{\spec}_j$. This alone does not make the problem easier to tackle. However, next, we will discuss how to generate application-specific parametric $\hat{\spec}_{j}$ and solve \cref{eq:individual-template} by optimizing over their parameters, allowing us to solve template generation much more efficiently than in \cref{eq:generation-opt}.

\subsection{Robustness to Adversarial Patches} \label{sec:robustn-advers-patch} %
We now instantiate the above scheme in order to verify the robustness of image classifiers against adversarial patches
\cite{ChiangNAZSG20}.  Consider an attacker that is allowed to
arbitrarily change any $p \times p$ patch of the image,
as showcased earlier in \cref{fig:patch}.  For such a
patch over pixel positions $([i,i+p-1] \times [j,j+p-1])$, the corresponding perturbation is
\begin{align*}
	&\spec_{p \times p}^{i,j}(\vx) := \{ \vz \in [0,1]^{h \times w} \mid \vz_{\pi_{i,j}^C}  = \vx_{\pi_{i,j}^C}  \}\\
  &\text{with}\; \pi_{i,j} = \left\{(k, l) \mid \begin{smallmatrix}k \in i, \dots, i+p-1\\l \in j, \dots, j+p-1\end{smallmatrix}  \right\}
\end{align*}
where $h$ and $w$ denote the height and width of the input $\vx$. Here $\pi_{i,j}$ denotes the parts of the image affected by the patch, and $\pi_{i,j}^C$ its complement, i.e., the unaffected part of the image.
To prove robustness for an arbitrarily placed $p \times p$
patch, however, one must consider the perturbation set $\spec_{p \times p}(\vx) := \cup_{i, j} \spec_{p \times p}^{i,j}(\vx)$.
\begin{figure}[ht]
\begin{minipage}{0.6\textwidth}
\begin{figure}[H]
  \def\mywidth{.32\textwidth}
  \begin{subfigure}{0.3\textwidth}
    \centering
\begin{tikzpicture}[scale=0.125]
\draw [step=1.0, draw=gray!50!white, fill=gray!15!white,  very thin] (0, 0) grid (10, 10) rectangle (0,0);
\node[text=my-full-blue] at (5,5) {$\mu_{1}$};
\end{tikzpicture}
    \subcaption{$\ell_{\infty}$}
  \end{subfigure}\hfill
  \begin{subfigure}{0.4\textwidth}
    \centering
\begin{tikzpicture}[scale=0.125]
\draw [step=1.0, draw=gray!50!white, fill=gray!15!white,  very thin] (0, 0) grid (2, 10) rectangle (0,0);

\draw [step=1.0, draw=gray!50!white, fill=gray!15!white,  very thin] (2, 0) grid (10, 2) rectangle (2,0);

\draw [step=1.0, draw=gray!50!white, fill=gray!15!white,  very thin] (2, 8) grid (10, 10) rectangle (2,8);

\draw [step=1.0, draw=gray!50!white, fill=gray!15!white,  very thin] (8, 2) grid (10, 8) rectangle (8,2);

\draw [step=1.0, draw=gray!50!white, fill=gray!35!white,  very thin] (2, 2) grid (8, 8) rectangle (2,2);

\node[text=my-full-blue] at (5,5) {$\mu_{1}$};
\node[text=my-full-orange] at (8.5,1) {$\mu_{2}$};
\end{tikzpicture}
    \subcaption{Center + Border}
  \end{subfigure}\hfill
  \begin{subfigure}{0.3\textwidth}
    \centering
\begin{tikzpicture}[scale=0.125]
\draw [step=1.0, draw=gray!50!white, fill=gray!15!white,  very thin] (0, 0) grid (5, 5) rectangle (0,0);
\draw [step=1.0, draw=gray!50!white, fill=gray!35!white,  very thin] (0, 5) grid (5, 10) rectangle (0,5);
\draw [step=1.0, draw=gray!50!white, fill=gray!15!white,  very thin] (5, 5) grid (10, 10) rectangle (5,5);
\draw [step=1.0, draw=gray!50!white, fill=gray!35!white,  very thin] (5, 0) grid (10, 5) rectangle (5, 0);

\node[text=my-full-blue] at (2.5,7.5) {$\mu_{1}$};
\node[text=my-full-orange] at (7.5,7.5) {$\mu_{2}$};
\node[text=my-full-red] at (2.5,2.5) {$\mu_{3}$};
\node[text=my-full-green] at (7.5,2.5) {$\mu_{4}$};
\end{tikzpicture}
    \subcaption{2x2 Grid}
  \end{subfigure}\\
\caption{Example splits $\mu$ for $10 \times 10$ pixels.} \label{fig:splits}
\end{figure}
\end{minipage}
\hfill
\begin{minipage}{0.38\textwidth}
\begin{figure}[H]
\vspace{0mm}
\hspace{6mm}\scalebox{0.4}{\begin{tikzpicture}
	\tikzset{>=latex}
	\node () [rectangle, minimum height=3cm, align=center, scale=0.7, rounded corners=2pt,
	anchor=center] at (11, 1) {
		\begin{tikzpicture}

			\fill [fill=my-full-gray] (7,4) -- (7,-3) -- (-5,-3) -- (-5, 4)  -- cycle;

			\coordinate (zono_head) at ({1.0000},{1.0000});
			\coordinate (zono_p_0) at ({-4.3000},{1.1000});
			\coordinate (zono_p_1) at ({-4.3000},{-1.1000});
			\coordinate (zono_p_2) at ({-3.9000},{-1.5000});
			\coordinate (zono_p_3) at ({-1.7000},{-1.5000});
			\coordinate (zono_p_4) at ({2.3000},{-0.7000});
			\coordinate (zono_p_5) at ({6.3000},{0.9000});
			\coordinate (zono_p_6) at ({6.3000},{3.1000});
			\coordinate (zono_p_7) at ({5.9000},{3.5000});
			\coordinate (zono_p_8) at ({3.7000},{3.5000});
			\coordinate (zono_p_9) at ({-0.3000},{2.7000});
			\fill [fill=my-blue, rounded corners=0mm] (zono_p_0) -- (zono_p_1) -- (zono_p_2) -- (zono_p_3) -- (zono_p_4) -- (zono_p_5) -- (zono_p_6) -- (zono_p_7) -- (zono_p_8) -- (zono_p_9)  -- cycle;

			\coordinate (box_head) at ({1.0000},{1.0000});
			\coordinate (box_p_0) at ({-4.3000},{-1.5000});
			\coordinate (box_p_1) at ({6.3000},{-1.5000});
			\coordinate (box_p_2) at ({6.3000},{3.5000});
			\coordinate (box_p_3) at ({-4.3000},{3.5000});
			\draw [draw=my-full-orange, line width=0.5mm, rounded corners=0mm] (box_p_0) -- (box_p_1) -- (box_p_2) -- (box_p_3)  -- cycle;

			\coordinate (box_ph_0) at ($(box_p_0) - (box_head)$);
			\coordinate (box_s_0) at ($(box_ph_0)!0.33!(0, 0) + (box_head)$);
			\coordinate (box_ph_1) at ($(box_p_1) - (box_head)$);
			\coordinate (box_s_1) at ($(box_ph_1)!0.33!(0, 0) + (box_head)$);
			\coordinate (box_ph_2) at ($(box_p_2) - (box_head)$);
			\coordinate (box_s_2) at ($(box_ph_2)!0.33!(0, 0) + (box_head)$);
			\coordinate (box_ph_3) at ($(box_p_3) - (box_head)$);
			\coordinate (box_s_3) at ($(box_ph_3)!0.33!(0, 0) + (box_head)$);
	
			\draw [color=my-full-green, line width=0.5mm, rounded corners=0mm] (box_s_0) -- (box_s_1) -- (box_s_2) -- (box_s_3)  -- cycle;

			\coordinate (dir) at ($(zono_p_9) - (zono_p_0)$);
			\coordinate (boundary0) at ($(box_s_3)+(dir)$);
			\coordinate (boundary1) at ($(box_s_3)-(dir)$);
			\draw [color=my-full-red, line width=0.5mm, rounded corners=0mm, dashed] ($(boundary0)!0.25!(box_s_3)$) -- ($(boundary1)!0.425!(box_s_3)$); 
			\node[text=my-full-blue] at (0.8, 0.85) {\Huge $N_{1:k}(\hat{\spec}_{i}(\vx, \eps_{i}))$};
			\node[text=my-full-orange] at (0.9, -2.2) {\Huge $T_k = \alpha_{\text{Box}}(N_{1:k}(\hat{\spec}_{i}(\vx, \eps_{i})))$};
			\node[text=my-full-green] at (5.4, -0.8) {\Huge $\beta_k T_k$};
		\end{tikzpicture}
		
	};

\end{tikzpicture}}
\vspace{-2.5mm}
\caption{Example Template.}
\label{fig:example_template}
\end{figure}
\end{minipage}
\end{figure}

To prove robustness for $\spec_{p \times p}$, existing approaches \cite{ChiangNAZSG20} separately verify $\spec_{p \times p}^{i,j}(\vx)$ for all
$i \in \{1, \dots, h-p+1\}, j \in \{1, \dots, w-p+1\}$.
For example, with $p=2$ and a $28 \times 28$ MNIST image, this approach requires 729 individual proofs.
Because the different proofs for $\spec_{p \times p}$ share similarities, this is an ideal candidate for proof sharing.
We utilize \cref{alg:verif_online} and check $\wedge_{i} v_{i}$ at the end to speed up this process.
For template generation, we solve \cref{eq:individual-template} for $m$ templates with an input perturbation $\hat{\spec}_{i}$ per template.

We empirically found that (recall \cref{tab:motivation}) setting $\hat{\spec}_{i}$ to an $\ell_{\infty}$ region $\spec_{\eps_i}$ to work particularly well to capture a majority of patch perturbations $\spec_{p \times p}^{i,j}$ at intermediate layers.
Specifically, we found that setting $\eps_i$ to the maximally verifiable value for this input to work particularly well.

To further increase the number of specifications contained in a set of templates $\mathcal{T}$,
we use $m$ template perturbations of the form
\begin{equation*}
  \hat{\spec}_{i}(\vx) := \{\vz \mid \| \vx_{\mu_{i}} - \vz_{\mu_{i}} \|_{\infty} \leq \epsilon_{i} \land \vx_{\mu_{i}^C} = \vz_{\mu_{i}^C} \},
\end{equation*}
where $\mu_{i}$ denotes a subset of pixels of the input image and $\mu_i^C$ its complement and we maximize $\eps_i$ in a best-effort manner.
In particular, we consider $\mu_{1}, \dots, \mu_{m}$, such that they partition the set of pixels in the image (e.g., in \cref{fig:splits}).

As noted earlier, this generation procedure needs to be fast, yet obtain $\mathcal{T}$ to which many abstractions match in order to obtain speed-ups.
Thus, we consider small $m$, and fixed patterns
$\mu_{1}, \dots, \mu_{m}$. For each $\hat{\spec}_{i}$, we aim to find the
largest $\epsilon_{i}$ which can still be verified in order to maximize the number of matches. Note that for $m=1$, this is equivalent to
the $\ell_{\infty}$ input perturbation $\spec_{\epsilon}$ with the maximally
verifiable $\epsilon$ for the given image.

Concretely, we can perform binary search over $\epsilon_{i}$ in order find a large $\eps_i$, still satisfying $N_{k+1:L}(\alpha_{\text{Box}}(N_{1:k}(\hat{\spec}_{i}))) \models \psi$.
Verification with our chosen DeepZ Zonotopes is not monotonous in $\eps_i$ due to the non-monotonic transformers used for non-linearities (e.g., ReLU).  This renders the application of binary search a best-effort approximation. As we don't require a formal maximum but rather aim to solve
a surrogate for  \cref{prob:generation}, this still works well in practice.
Further note that, applying $\alpha_{\text{Box}}$ to templates introduces imprecision, i.e. $V_T$ might not be able to prove properties over templates that it could without the application of $\alpha_{\text{Box}}$.
However, \cref{thm:precission} (which only requires properties of $V_S$) still applies.

\begin{wrapfigure}[20]{R}{0.53\textwidth}
\vspace{-2.5em}
\centering
\begin{minipage}{0.52\textwidth}
\begin{algorithm}[H]
\SetInd{0.425em}{0.4em}
\DontPrintSemicolon
\SetAlgoLined
\SetNoFillComment
\SetKwInOut{KwInput}{Input}
\KwInput{$\vx, N, \mu_{1}, \dots, \mu_{m}, K, \psi, V_T$}
\KwResult{$\mathcal{T}^{k}$ for $k \in K$}
$\mathcal{T}^{k} \gets \{\}$ for $k \in K$\;
\For{$i \gets 1$ \KwTo $m$}{
  $\hat{\spec}_{i}(\vx, \eps) := \{ \vz \mid \| \vx_{\mu_{i}} - \vz_{\mu_{i}} \|  \leq \epsilon$\\\qquad\qquad\qquad\quad$\land\; \vx_{\mu_{i}^C} = \vz_{\mu_{i}^C}  \} $\;
  $f(\eps) := V_T(\hat{\spec}_{i}(\vx, \eps), N) \models \psi $ \;
  $\eps_{i} \gets $ \texttt{bin\_search}$(\eps, f(\eps))$\;  \nllabel{alg:online_generation:search}
\For{$k \in K$}{
    $T_k \gets \alpha_{\text{Box}}( V_T(\hat{\spec}_{i}(\vx, \eps_{i}), N_{1:k}) )$\;
    $g(\beta_k) := V_T(\beta T_k, N_{k+1:L}) \models \psi $\;
	$\beta_{k} \gets $ \texttt{bin\_search}$(\beta, g(\beta))$\;\nllabel{alg:online_generation:search2}
	$\mathcal{T}^{k} \gets \mathcal{T}^{k} \cup \{\beta_k T_k\}$\;
  }
 }
 \Return{$\mathcal{T}^{k}$ for $k \in K$}
 \caption{Online Template Generation for Patches}
 \label{alg:online_generation}
\end{algorithm}
\end{minipage}
\end{wrapfigure}

\paragraph{Templates at multiple layers} %
We can extend this approach to obtain templates at multiple layers without a large increase in computational cost.
With templates at multiple layers, we first try to match the propagated shape against the earliest template layer and upon failure propagate it further to the next, where we again attempt to match the template.
In \cref{alg:verif_online}, this means repeating the block from Line~\ref{alg:verif_online:match_start}~to~Line~\ref{alg:verif_online:match_end} for each template layer before going on to the check on \cref{alg:verif_online:check}.

The full template generation procedure is given in \cref{alg:online_generation}.
First, we perform a binary search over $\eps_i$ (\cref{alg:online_generation:search}) to find the largest $\eps_i$, for which the specification is verifiable.
Then for each layer $k$ in the set of layers $K$ at which we are creating templates we create a box $T_k$ from the Zonotope. As this $T_k$ may not be verifiable, due to the imprecision added in $\alpha_{\text{Box}}$, we then perform another binary search for the largest scaling factor $\beta_k$ (\cref{alg:online_generation:search2}), which is applied to the matrix $\mA$ in \cref{eq:zonotope}. We denote this operation as $\beta_k T_k$.
We show an example for a single layer $k$ in \cref{fig:example_template}. The blue area outlines the Zonotope found via \cref{alg:online_generation:search}, which is verifiable as it is fully on one side of the decision boundary (red, dashed). After applying $\alpha_{\text{Box}}$ (orange), however, is not (crosses the decision boundary). By scaling it with $\beta_k$ the shape is verifiable again (green) and used as a template.

\subsection{Geometric Robustness}\label{sec:geometric-robustness} %
Geometric robustness verification
\cite{PeiCYJ17,DeepPoly,DeepG,fischer2020transformationsmoothing}
aims to verify the robustness of neural networks against geometric transformations such as image rotations
or translations. These transformations typically include an interpolation
operation.
For example consider rotation $R_{\gamma}$ of an image by
$\gamma \in \Gamma$ degrees for an interval $\Gamma$ (e.g.,
$\gamma \in [-5, 5]$), for which we consider the specification 
$\spec_{\Gamma}(\vx) := \{R_{\gamma}(\vx) \mid \gamma \in
\Gamma \}$.
We note that, unlike $\ell_\infty$ and patch verification, the input
regions for geometric transformations are non-linear and have no
closed-form solutions. Thus, an overapproximation of the input region
must be obtained \cite{DeepG}.  For large $\Gamma$, the approximate input
region $\spec_{\Gamma}(\vx)$, can be too coarse resulting in
imprecise verification.  Hence, in order to assert $\psi$ on
$\spec_{\Gamma}$, existing state-of-the-art approaches \cite{DeepG}, split $\Gamma$ into $r$ smaller ranges
$\Gamma_{1}, \dots, \Gamma_{r}$ and then verify the resulting $r$ specifications 
$(\spec_{\Gamma_{i}}, \psi)$ for $i \in 1, \dots, r$.
These smaller perturbations share similarities facilitating proof
sharing. We instantiate our approach similar to
\cref{sec:robustn-advers-patch}.  A key difference to
\cref{sec:robustn-advers-patch} is that while
$\vx \in \spec_{p \times p}^{i,j}(\vx)$ for all $i, j$ in patches, here in
general $\vx \not\in \spec_{\Gamma_{i}}(\vx)$ for most $i$. Therefore,
the individual perturbations $\spec_{i}(\vx)$ do not overlap.
To account for this, we consider $m$ templates and split $\Gamma$ into
$m$ equally sized chunks (unrelated to the $r$ splits) obtaining the
angles $\gamma_{1}, \dots, \gamma_{m}$ at the center of each chunk.
For $m$ templates we then consider the perturbations
$\hat{\spec}_{i} := \spec_{\epsilon_{i}}(R_{\gamma_{i}}(\vx))$,
denoting the $\ell_{\infty}$ perturbation of size $\eps_{i}$ around the $\gamma_{i}$ degree rotated $\vx$. To find the template we employ a procedure analogous to \cref{alg:online_generation}.

\subsection{Requirements for Proof Sharing} \label{sec:requirements-for-proof-sharing} %
Now, we discuss the requirements on the neural network $N$ such that proof sharing via templates works well.
For simplicity, we discuss simple per-dimension box bounds propagation for $V_S$ and $V_T$. However, similar arguments can be made for more complex relational abstractions such as Zonotopes or Polyhedra.

In order for an abstraction $S$ to match to a template $T$, we need to show interval inclusion for each dimension.
For a particular dimension $i$  this can occur in two ways:
(i) when both $S$ and $T$ are just a point in that dimension and these points coincide, e.g., $a^S_i = a^T_i$, or
(ii) when $a^S_i \pm d^S_i \subseteq a^T_i \pm d^T_i$.
While particularly in ReLU networks, the first case can occur after a ReLU layer sets values to zero, we focus our analysis here on the second case as it is more common. In this case, the width of $T$ in that dimension $d^T_i$ must be sufficient to cover $S$.
Ignoring case (i) and letting $\supp(T)$ denote the dimensions in which $d_i^T > 0$, we can pose that $\supp(S) \subseteq \supp(T)$ as a necessary condition for inclusion.
While it is in general hard to argue about the magnitudes of these values, this approach still provides an intuition.
When starting from  input specifications $\supp(\spec) \not\subseteq \supp(\hat{\spec})$,
$\supp(S) \subseteq \supp(T)$ can only occur if during propagation through the neural network $N_{1:k}$
the mass in $\supp(\hat{\spec})$ can "spread out" sufficiently to cover $\supp(S)$. 
In the fully connected neural networks that we discuss here, the matrices of linear layers provide this possibility.
However, in networks that only read part of the input at a time such as recurrent neural networks, or convolutional neural networks in which only locally neighboring inputs feed into the respective output in the next layer, these connections do not necessarily exist. This makes proof sharing hard until layers later in the neural network, that regionally or globally pool information.
As this increases the depth of the layer $k$ at which proof transfer can be applied, this also decreases the potential speed-up of proof transfer.
This could be alleviated by different ways of creating templates, which we plan to investigate in the future.

\section{Experimental Evaluation}\label{sec:evaluation} %
We now experimentally evaluate the effectiveness of our algorithms from \cref{sec:template-generation-individual}.

\subsection{Experimental Setup} \label{sec:experimental-details} %
\label{sec:details-motivation}
We consider the verification of robustness to adversarial patch attacks and geometric transformations in \cref{sec:robustn-agsinst-adve} and \cref{sec:robustn-against-geom}, respectively. We define specifications on the first 100 test set images each from the MNIST \cite{MNIST} and the CIFAR-10 dataset \cite{cifar} ("CIFAR") as with repetitions and parameter variations the overall runtime becomes high. We use DeepZ \cite{SinghGMPV18} as the baseline verifier as well as for $V_S$ and $V_T$ \cite{SinghGMPV18}.
Throughout this section, we evaluate proof sharing for two networks on two common datasets: We use a seven layer neural network with 200 neurons per layer ("7x200") and a nine layer network with 500 neurons per layer ("9x500") for both the MNIST\cite{MNIST} and CIFAR datasets \cite{cifar}, both utilizing ReLU activations. 
These architectures are similar to the fully-connected ones used in the ERAN and Mnistfc VNN-Comp categories \cite{VNN21}.
\setlength{\tabcolsep}{0.5em}
\begin{table}[t]
    \centering
	\caption{Rate of $\spec_{2 \times 2}^{i,j}$ matched to templates $\mathcal{T}$ for $\spec_{2 \times 2}$ patch verification for different combinations of template layers $k$, 7x200 networks,using $m=1$
      template.}
    \label{tab:patches_layers_matches}
	    \vspace{2mm}
    \begin{tabular}{@{}lcrrrrrrrcc@{}}\toprule
   \multicolumn{2}{r}{template at layer $k$}& 1 & 2 & 3 & 4 & 5 & 6 & 7 && patch verif. [\%]\\
    \midrule
   MNIST && 18.6 & 85.6 & 94.1 & 95.2 & 95.5 & 95.7 & 95.7 && 97.0\\
    CIFAR && 0.1 & 27.1 & 33.7 & 34.4 & 34.2 & 34.2 & 34.3 && 42.2\\
    \bottomrule
    \end{tabular}
\end{table}

\setlength{\tabcolsep}{0.5em}
\begin{table}[t]
    \centering
          \caption{Average verification time in seconds per image for $\spec_{2 \times 2}$ patchs for different combinations of template layers $k$, 7x200 networks,using $m=1$ template.}
    \label{tab:patches_layers_timings}
	    \vspace{2mm}
    \begin{tabular}{@{}lcrrrrrrrrrr@{}}\toprule
   & & &\multicolumn{8}{c}{Proof Sharing, template layer $k$}\\
    \cmidrule{4-11}
    & & Baseline & 1 & 2 & 3 & 4 & 1+3 & 2+3 & 2+4 & 2+3+4\\
    \midrule
    MNIST && 2.10 & 1.94 & 1.15 & 1.22 & 1.41 & 1.27 & \textbf{1.09} & 1.10 & 1.14\\
    CIFAR && 3.27 & 2.98 & 2.53 & 2.32 & 2.47 & 2.35 & 2.49 & \textbf{2.42} & 2.55\\
    \bottomrule
    \end{tabular}
\end{table}

For MNIST, we train 100 epochs, enumerating all patch locations for each sample, and for CIFAR we train for 600 with 10 random patch locations, as outlined in \cite{ChiangNAZSG20} with interval training \cite{DiffAI,IBP}. On MNIST the 7x200 and the 9x500 achieve a natural accuracy of of 98.3\% and 95.3\% respectively. For CIFAR, these values are 48.8\% and 48.1\% respectively. 
Our implementation utilizes PyTorch \cite{PyTorch} and is evaluated on Ubuntu 18.04 with an Intel Core i9-9900K CPU and 64 GB RAM. For all timing results, we provide the mean over three runs.

\subsection{Robustness against adversarial patches}\label{sec:robustn-agsinst-adve} %
For MNIST, containing $28 \times 28$ images, as outlined
in \cref{sec:robustn-advers-patch}, in order to verify inputs to be
robust against $2 \times 2$ patch perturbations, 729 individual
perturbations must be verified.  Only if all are verified,
the overall property can be verified for a given image.  Similarly, for
CIFAR, containing $32 \times 32$ color images, there are 961 individual perturbations (the patch is applied over all color channels).

\setlength{\tabcolsep}{0.5em}
\begin{table}[ht]
    \centering
       \caption{$\spec_{2 \times 2} $ patch verification with templates at the
        \nth{2}~\&~\nth{3} layer of the 7x200 networks for different masks.}
    \label{tab:patches_template_masks}
	\vspace{2mm}
   \begin{tabular}{@{}lccc@{}}\toprule
    Method/Mask & m & patch matched [\%] & $t$ [s]\\
    \midrule
    Baseline & - & - & 2.14\\
    L-infinity & 1 & 94.1 & \textbf{1.11} \\
    Center + Border & 2 & 94.6 & 1.41\\
    2x2 Grid & 4 & \textbf{95.0} & 3.49 \\
    \bottomrule
    \end{tabular}
\end{table}

\setlength{\tabcolsep}{0.5em}
\begin{table}[ht]
\centering
          \caption{$\spec_{2 \times 2} $ patch verification with templates
        generated on the second and third layer using the
        $\ell_\infty$-mask. Verification times are given for the
        baseline $t^{BL}$ and for applying proof sharing $t^{PS}$ in seconds per image.}
    \label{tab:patches_nets}
	\vspace{2mm}
    \begin{tabular}{@{}llcrcrrccc@{}}\toprule
      Dataset&
      Net&
      verif. acc. [\%] &&
      $t^{BL}$ &
      $t^{PS}$ &&
      patch mat. [\%] &
      patch verif. [\%]\\
    \midrule
    \multirow{2}{*}{MNIST} & 7x200 & 81.0 && 2.10 & \textbf{1.10} && 94.1 & 97.0\\
     & 9x500 & 66.0 && 2.70 & \textbf{1.32} && 93.0 & 95.3\\
    \midrule
      \multirow{2}{*}{CIFAR} & 7x200 & 29.0 && 3.28 & \textbf{2.45} && 33.7 & 42.2\\
    & 9x500 & 28.0 && 5.48  & \textbf{4.48} && 34.2 & 46.2\\
    \bottomrule
    \end{tabular}
\end{table}

We now investigate the two main parameters of \cref{alg:online_generation}:
the masks $\mu_{1}, \dots, \mu_{m}$ and the layers $k \in K$. We first study the impact of the layer $k$ used for creating the template. To this end, we consider
the 7x200 networks, use $m=1$ (covering the whole
image; equivalent to $\hat{\spec}_{\epsilon}$).  \cref{tab:patches_layers_matches} shows the corresponding template matching
rates, and the overall percentage of individual patches that can be verified ``patches verif.''.
(The overall percentage of images for which $\spec_{2\times2}$ is true is reported  as ``verf.'' in \cref{tab:patches_nets}.)
\cref{tab:patches_layers_timings} shows the corresponding verification times (including the template generation).  We observe that many template matches can
already be made at the second or third layer.  
As creating templates simultaneously at the second and third layer works well for both datasets, we utilize templates at these layers in further experiments.

Next, we investigate the impact of the pixel masks
$\mu_{1}, \dots, \mu_{m}$. To this end, we consider three different
settings, as showcased in \cref{fig:splits} earlier: (i) the full image ($\ell_{\infty}$-mask as before; $m=1$), (ii) "center + border" ($m=2$), where we consider the $6 \times 6$ center pixel as one group and all others as another, and (iii) the $2 \times 2$ grid
($m = 4$) where we split the image into equally sized quarters. %

As we can see in \cref{tab:patches_template_masks}, for higher $m$ more patches can be matched to the
templates, indicating that our optimization procedure is a good approximation to \cref{prob:generation}, which only considers the number of templates matched.
Yet, for $m>1$ the increase in matching rate $p$ does not offset the additional time in template generation and matching. Thus, $m=1$ results in a better trade-off.
This result highlights the trade-offs discussed throughout sections \cref{sec:transfer,sec:template-generation-individual}.
Based on this investigation we now, in \cref{tab:patches_nets},
evaluate all networks and datasets using $m=1$ and template generation at
layers 2 and 3.  In all cases, we obtain a speed up between
$1.2$ to $2 \times$ over the baseline verifier.
Going from $2 \times 2$ to $3 \times 3$ patches speed ups remain around 1.6 and 1.3 for the two datasets respectively.

\setlength{\tabcolsep}{0.5em}
\begin{table}[ht]
\centering
    \caption{Speed-ups achievable in the setting of \cref{tab:patches_layers_matches}. $t^{BL}$ the baseline.
    }
	\vspace{2mm}
    \label{tab:optimal_speedup}
    \begin{tabular}{@{}lrrrrr@{}}\toprule
      & & \multicolumn{4}{c}{speedup at layer $k$}\\
      \cmidrule{3-6}
Layer $k$& & 1 & 2 & 3 & 4\\
\midrule
realized & $t^{BL} / t^{PS}$ & 1.08  & 1.83 & 1.72 & 1.49\\
\midrule
optimal & $t^{BL} / (t_T + r t_S + r t_{\subseteq})$ & 3.75 & 2.51 & 1.92 & 1.56\\
optimal, no $\subseteq$ & $ t^{BL} /(t_T + rt_S)$ & 4.02 & 2.68 & 2.01 & 1.62\\
optimal, no gen $\mathcal{T}$., no $\subseteq$ & $t^{BL} / rt_S $  & 4.57 & 2.90 & 2.13 & 1.69\\
   \bottomrule
    \end{tabular}
\end{table}

\paragraph{Comparison with theoretically achievable speed-up}
Finally, we want to determine the maximal possible speed-up with proof sharing and see how much of this potential is realized by our method.
To this end we investigate the same setting and network as in \cref{tab:patches_layers_matches}.
We let $t^{BL}$ and $t^{PS}$ denote the runtime of the base verifier without and with proof sharing respectively. Similar to the discussion in \cref{sec:template-generation-individual} we can break down $t^{PS}$ into $t_T$ (template generation time), $t_S$ (time to propagate one input to layer $k$), $t_{\subseteq}$ (time to perform template matching) and $t_\post$ (time to verify $S$ if no match).
\cref{tab:optimal_speedup} shows different ratios of these quantities. For all, we assume a perfect matching rate at layer $k$ and calculate the achievable speed-up for patch verification on MNIST. Comparing the optimal and realized results, we see that at layers 3 and 4 our template generation algorithm, despite only approximately solving \cref{prob:generation} achieves near-optimal speed-up.
By removing the time for template matching and template generation we can see that, at deeper layers, speeding up $t_{\subseteq}$ and $t_{\mathcal{T}}$ only yield diminishing returns.

\subsection{Robustness against geometric perturbations}\label{sec:robustn-against-geom} %
For the verification of geometric perturbations, we take 100 images from the
MNIST dataset and the 7x200 neural network from
\cref{sec:robustn-agsinst-adve}.
In \cref{tab:geometric_splits}, we consider an input region
with $\pm$2\textdegree~rotation, $\pm$10\% contrast and $\pm$1\% brightness change, inspired by \cite{DeepG}.
To verify this region,  similar to existing approaches \cite{DeepG}, we choose to split the rotation into $r$ regions, each yielding a Box specification over the input.
Here we use $m=1$, a single template, with the largest verifiable $\epsilon$ found via binary search. We observe that as we
increase $r$, the verification rate increases, but also the
speed ups. Proof sharing enables significant speed-up between $1.6$ to $2.9 \times$.

\setlength{\tabcolsep}{0.5em}
\begin{table}[ht]
    \centering
        \caption{ $\pm$2\textdegree\ rotation, $\pm$10\% contrast and $\pm$1\%
        brightness change split into $r$ perturbations on 100
        MNIST images.  Verification rate, rate of splits matched and verified along with the run time of Zonotope $t^{BL}$
        and proof sharing $t^{PS}$.}
    \label{tab:geometric_splits}
	\vspace{2mm}

    \begin{tabular}{@{}ccccrr@{}}\toprule

	    $r$ & verif. [\%] & splits verif. [\%] & splits matched [\%] & $t^{BL}$ & $t^{PT}$ \\
    \midrule
    4 & 73.0 & 87.3 & 73.1 & 3.06 & \textbf{1.87}\\
    6 & 91.0 & 94.8 & 91.0  & 9.29 & \textbf{3.81}\\
    8 & 93.0 & 95.9 & 94.2 & 20.64 & \textbf{7.48}\\
    10 & 95.0 & 96.5 & 94.9 & 38.50 & \textbf{13.38}\\
    \bottomrule
    \end{tabular}
\end{table}

\setlength{\tabcolsep}{0.5em}
\begin{table}[t]
    \centering
      \caption{$\pm$40\textdegree\ rotation  split into 200 perturbations
        evaluated on MNIST. The verification rate is
        just 15 \%, but 82.1 \% of individual splits can be verified.}
    \label{tab:geometric_multi_templates}
	    \vspace{2mm}
    \begin{tabular}{@{}lccc@{}}\toprule
    Method & $m$ & splits matched [\%] & $t$ [s]\\
    \midrule
      
      Baseline & - & - & 11.79\\
      \midrule
    \multirow{3}{*}{Proof Sharing} & 1 & 38.0 & 9.15\\
     & 2 & 41.1 & 9.21\\
     & 3 & 58.5 & \textbf{8.34}\\
    \bottomrule
    \end{tabular}
\end{table}

Finally, we investigate the impact of the number of templates $m$.  
To this end, we consider a setting with a large parameter space:
$\pm$40\textdegree~rotation generated input region with $r=200$.  In \cref{tab:geometric_multi_templates}, we evaluate this
for $m$ templates obtained from the $\ell_{\infty}$ input
perturbation around $m$ equally spaced rotations, where we apply binary search to find $\epsilon_i$ tailored for each template.  Again we observe
that $m>1$ allows more templates matches.
However, in this setting the relative increase is much larger than for patches, thus making $m=3$ faster than $m=1$.

\subsection{Discussion}
\label{sec:discussion}
We have shown that proof sharing can achieve speed-ups over conventional execution.
While the speed-up analysis (see \cref{sec:template-generation-individual} and \cref{tab:optimal_speedup}) put a ceiling on what is achievable in particular settings,  we are optimistic that proof sharing can be an important tool for neural network robustness analysis.
In particular, as the size of certifiable neural networks continues to grow, the potential for gains via proof sharing is equally growing.
Further, the idea of proof effort reuse can enable efficient verification of larger disjunctive specifications such as the patch or geometric examples considered here. 
Besides the immediately useful speed-ups, the concept of proof sharing is interesting in its own right and can provide insights into the learning mechanisms of neural networks.

\section{Related Work}\label{sec:related-work} %
Here, we briefly discuss conceptually related work:

\paragraph{Incremental Model Checking}
The field of model checking aims to show whether a formalized model, e.g. of software or hardware, adheres to a specification. As neural network verification can also be cast as model checking, we review incremental model checking techniques which utilize a similar idea to proof sharing: reuse partial previous computations when checking new models or specifications. 
Proof sharing has been applied for discovering and reusing lemmas when proving theorems 
for satisfiability \cite{Bradley11}, Linear Temporal Logic \cite{BradleySHZ11}, and modal $\mu$-calculus \cite{SokolskyS94}.
Similarly, caching solvers \cite{TaljaardGV20} for Satisfiability Modulo
Theories cache obtained results or even the full models used to
obtain the solution, with assignments for all variables, allowing for
faster verification of subsequent queries. 
For program analysis tasks that deal with repeated similar inputs (e.g. individual commits in a software project) can leverage partial results \cite{YangDR09}, constraints
\cite{VisserGD12} precision information \cite{BeyerLNSW13,BeyerW13} from previous runs.

\paragraph{Proof Sharing Between Networks}
In neural network verification, some approaches abstract the network to achieve speed-ups in verification. These simplifications are constructed in a way that the proof can be adapted for the original neural network \cite{AshokHKM20,ZhongTLZK21}.
Similarly, another family of approaches analyzes the difference between
two closely related neural networks by utilizing their structural similarity \cite{PaulsenWW20,PaulsenWWW20}. Such approaches can be used to reuse analysis results between neural network modifications, e.g. fine-tuning \cite{ChengY20,WeiL21}.

In contrast to these works, we do not modify the neural network, but achieve speed-ups rather by only considering the relaxations obtained in the proofs.
\cite{WeiL21} additionally consider small changes to the input, however, these are much smaller than the difference in specification we consider here.

\section{Conclusion}\label{sec:conclusion} %
We introduced the novel concept of proof sharing in the context of neural network verification.
We showed how to instantiate this concept, achieving speed-ups of up to $2$ to $3$ x
for patch verification and geometric verification.
We believe that the ideas introduced in this work can serve as a solid foundation for exploring methods that effectively share proofs in neural network verification.

\bibliographystyle{splncs04}
\bibliography{references}

\begin{thebibliography}{10}
\providecommand{\url}[1]{\texttt{#1}}
\providecommand{\urlprefix}{URL }
\providecommand{\doi}[1]{https://doi.org/#1}

\bibitem{AshokHKM20}
Ashok, P., Hashemi, V., Kret{\'{\i}}nsk{\'{y}}, J., Mohr, S.: Deepabstract:
  Neural network abstraction for accelerating verification. In: Proc of.
  Automated Technology for Verification and Analysis (ATVA). vol. 12302 (2020)

\bibitem{bak2017simulation}
Bak, S., Duggirala, P.S.: Simulation-equivalent reachability of large linear
  systems with inputs. In: Proc. of Computer Aided Verification (CAV). Springer
  (2017)

\bibitem{VNN21}
Bak, S., Liu, C., Johnson, T.T.: The second international verification of
  neural networks competitio. ArXiv preprint  \textbf{abs/2109.00498} (2021)

\bibitem{DeepG}
Balunovic, M., Baader, M., Singh, G., Gehr, T., Vechev, M.T.: Certifying
  geometric robustness of neural networks. In: Neural Information Processing
  Systems (NIPS) (2019)

\bibitem{BeyerLNSW13}
Beyer, D., L{\"{o}}we, S., Novikov, E., Stahlbauer, A., Wendler, P.: Precision
  reuse for efficient regression verification. In: Symposium on the Foundations
  of Software Engineering (SIGSOFT) (2013)

\bibitem{BeyerW13}
Beyer, D., Wendler, P.: Reuse of verification results - conditional model
  checking, precision reuse, and verification witnesses. In: Proc. of Model
  Checking Software. vol.~7976 (2013)

\bibitem{Bradley11}
Bradley, A.R.: Sat-based model checking without unrolling. In: Proc. of
  Verification, Model Checking, and Abstract Interpretation (VMCAI). vol.~6538
  (2011)

\bibitem{BradleySHZ11}
Bradley, A.R., Somenzi, F., Hassan, Z., Zhang, Y.: An incremental approach to
  model checking progress properties. In: International Conf. on Formal Methods
  in Computer-Aided Design (FMCAD) (2011)

\bibitem{Brown17Patches}
Brown, T.B., Man{\'{e}}, D., Roy, A., Abadi, M., Gilmer, J.: Adversarial patch.
  ArXiv preprint  \textbf{abs/1712.09665} (2017)

\bibitem{ChengY20}
Cheng, C., Yan, R.: Continuous safety verification of neural networks. In:
  Design, Automation {\&} Test in Europe Conf. {\&} Exhibition (2021)

\bibitem{ChiangNAZSG20}
Chiang, P., Ni, R., Abdelkader, A., Zhu, C., Studer, C., Goldstein, T.:
  Certified defenses for adversarial patches. In: Proc. of International Conf.
  on Learning Representations (ICLR) (2020)

\bibitem{Cousot:77}
Cousot, P., Cousot, R.: Abstract interpretation: A unified lattice model for
  static analysis of programs by construction or approximation of fixpoints.
  In: Proc. of Principles of Programming Languages (POPL) (1977)

\bibitem{Cousot:78}
Cousot, P., Halbwachs, N.: Automatic discovery of linear restraints among
  variables of a program. In: Proc. of Principles of Programming Languages
  (POPL) (1978)

\bibitem{fischer2020transformationsmoothing}
Fischer, M., Baader, M., Vechev, M.T.: Certified defense to image
  transformations via randomized smoothing. In: Neural Information Processing
  Systems (NIPS) (2020)

\bibitem{fischer2022shared}
Fischer, M., Sprecher, C., Dimitrov, D.I., Singh, G., Vechev, M.: Shared
  certificates for neural network verification. In: Computer Aided
  Verification: 34th International Conference, CAV 2022, Haifa, Israel, August
  7--10, 2022, Proceedings, Part I. pp. 127--148. Springer (2022)

\bibitem{AI2}
Gehr, T., Mirman, M., Drachsler{-}Cohen, D., Tsankov, P., Chaudhuri, S.,
  Vechev, M.T.: {AI2:} safety and robustness certification of neural networks
  with abstract interpretation. In: Symposium on Security and Privacy (S\&P)
  (2018)

\bibitem{GoubaultP15}
Goubault, E., Putot, S.: A zonotopic framework for functional abstractions.
  Formal Methods Syst. Des.  \textbf{47}(3) (2015)

\bibitem{IBP}
Gowal, S., Dvijotham, K., Stanforth, R., Bunel, R., Qin, C., Uesato, J.,
  Arandjelovic, R., Mann, T.A., Kohli, P.: On the effectiveness of interval
  bound propagation for training verifiably robust models. ArXiv preprint
  \textbf{abs/1810.12715} (2018)

\bibitem{Reluplex}
Katz, G., Barrett, C.W., Dill, D.L., Julian, K., Kochenderfer, M.J.: Reluplex:
  An efficient {SMT} solver for verifying deep neural networks. In: Proc. of
  Computer Aided Verification (CAV). vol. 10426 (2017)

\bibitem{Marabou:19}
Katz, G., Huang, D.A., Ibeling, D., Julian, K., Lazarus, C., Lim, R., Shah, P.,
  Thakoor, S., Wu, H., Zelji{\'{c}}, A., Dill, D.L., Kochenderfer, M.J.,
  Barrett, C.: The marabou framework for verification and analysis of deep
  neural networks. In: Proc. of Computer Aided Verification (CAV) (2019)

\bibitem{KopetzkiSA17}
Kopetzki, A., Sch{\"{u}}rmann, B., Althoff, M.: Methods for order reduction of
  zonotopes. In: Conf. on Decision and Control (CDC) (2017)

\bibitem{cifar}
Krizhevsky, A., Hinton, G., et~al.: Learning multiple layers of features from
  tiny images  (2009)

\bibitem{KrizhevskySH17}
Krizhevsky, A., Sutskever, I., Hinton, G.E.: Imagenet classification with deep
  convolutional neural networks. In: Neural Information Processing Systems
  (NIPS) (2012)

\bibitem{MNIST}
LeCun, Y., Boser, B.E., Denker, J.S., Henderson, D., Howard, R.E., Hubbard,
  W.E., Jackel, L.D.: Handwritten digit recognition with a back-propagation
  network. In: Neural Information Processing Systems (NIPS) (1989)

\bibitem{MadryTraining}
Madry, A., Makelov, A., Schmidt, L., Tsipras, D., Vladu, A.: Towards deep
  learning models resistant to adversarial attacks. In: Proc. of International
  Conf. on Learning Representations (ICLR) (2018)

\bibitem{DiffAI}
Mirman, M., Gehr, T., Vechev, M.T.: Differentiable abstract interpretation for
  provably robust neural networks. In: Proc. of International Conf. on Machine
  Learning (ICML). vol.~80 (2018)

\bibitem{PyTorch}
Paszke, A., Gross, S., Massa, F., Lerer, A., Bradbury, J., Chanan, G., Killeen,
  T., Lin, Z., Gimelshein, N., Antiga, L., Desmaison, A., K{\"{o}}pf, A., Yang,
  E., DeVito, Z., Raison, M., Tejani, A., Chilamkurthy, S., Steiner, B., Fang,
  L., Bai, J., Chintala, S.: Pytorch: An imperative style, high-performance
  deep learning library. In: Neural Information Processing Systems (NIPS)
  (2019)

\bibitem{PaulsenWW20}
Paulsen, B., Wang, J., Wang, C.: Reludiff: differential verification of deep
  neural networks. In: International Conf. on Software Engineering (ICSE)
  (2020)

\bibitem{PaulsenWWW20}
Paulsen, B., Wang, J., Wang, J., Wang, C.: {NEURODIFF:} scalable differential
  verification of neural networks using fine-grained approximation. In: Conf.
  on Automated Software Engineering (ASE) (2020)

\bibitem{PeiCYJ17}
Pei, K., Cao, Y., Yang, J., Jana, S.: Towards practical verification of machine
  learning: Th. ArXiv preprint  \textbf{abs/1712.01785} (2017)

\bibitem{RothLKB03}
Roth, V., Laub, J., Kawanabe, M., Buhmann, J.M.: Optimal cluster preserving
  embedding of nonmetric proximity data. {IEEE} Trans. Pattern Anal. Mach.
  Intell.  \textbf{25}(12) (2003)

\bibitem{SadraddiniT19}
Sadraddini, S., Tedrake, R.: Linear encodings for polytope containment
  problems. In: Conf. on Decision and Control (CDC) (2019)

\bibitem{AlphaGo}
Silver, D., Schrittwieser, J., Simonyan, K., Antonoglou, I., Huang, A., Guez,
  A., Hubert, T., Baker, L., Lai, M., Bolton, A., et~al.: Mastering the game of
  go without human knowledge. Nature  \textbf{550}(7676) (2017)

\bibitem{SinghGMPV18}
Singh, G., Gehr, T., Mirman, M., P{\"{u}}schel, M., Vechev, M.T.: Fast and
  effective robustness certification. In: Neural Information Processing Systems
  (NIPS) (2018)

\bibitem{DeepPoly}
Singh, G., Gehr, T., P{\"{u}}schel, M., Vechev, M.T.: An abstract domain for
  certifying neural networks. {PACMPL}  \textbf{3}({POPL}) (2019)

\bibitem{SokolskyS94}
Sokolsky, O., Smolka, S.A.: Incremental model checking in the modal
  mu-calculus. In: Proc. of Computer Aided Verification (CAV). vol.~818 (1994)

\bibitem{SzegedyIntriguing}
Szegedy, C., Zaremba, W., Sutskever, I., Bruna, J., Erhan, D., Goodfellow,
  I.J., Fergus, R.: Intriguing properties of neural networks. In: Proc. of
  International Conf. on Learning Representations (ICLR) (2014)

\bibitem{TaljaardGV20}
Taljaard, J., Geldenhuys, J., Visser, W.: Constraint caching revisited. In:
  Proc. of {NASA} Formal Methods (NFM). vol. 12229 (2020)

\bibitem{TjengMILP}
Tjeng, V., Xiao, K.Y., Tedrake, R.: Evaluating robustness of neural networks
  with mixed integer programming. In: Proc. of International Conf. on Learning
  Representations (ICLR) (2019)

\bibitem{tran2019star}
Tran, H.D., Lopez, D.M., Musau, P., Yang, X., Nguyen, L.V., Xiang, W., Johnson,
  T.T.: Star-based reachability analysis of deep neural networks. In:
  International Symposium on Formal Methods (FME). Springer (2019)

\bibitem{VisserGD12}
Visser, W., Geldenhuys, J., Dwyer, M.B.: Green: reducing, reusing and recycling
  constraints in program analysis. In: Symposium on the Foundations of Software
  Engineering (SIGSOFT) (2012)

\bibitem{WeiL21}
Wei, T., Liu, C.: Online verification of deep neural networks under domain or
  weight shift. ArXiv preprint  \textbf{abs/2106.12732} (2021)

\bibitem{Weng2018}
Weng, T., Zhang, H., Chen, H., Song, Z., Hsieh, C., Daniel, L., Boning, D.S.,
  Dhillon, I.S.: Towards fast computation of certified robustness for relu
  networks. In: Proc. of International Conf. on Machine Learning (ICML).
  vol.~80 (2018)

\bibitem{trustworthyacm}
Wing, J.M.: Trustworthy ai. Commun. ACM  \textbf{64}(10) (2021)

\bibitem{WongK18}
Wong, E., Kolter, J.Z.: Provable defenses against adversarial examples via the
  convex outer adversarial polytope. In: Proc. of International Conf. on
  Machine Learning (ICML). vol.~80 (2018)

\bibitem{YangDR09}
Yang, G., Dwyer, M.B., Rothermel, G.: Regression model checking. In:
  International Conf. on Software Maintenance {(ICSM)} (2009)

\bibitem{Zhang2018}
Zhang, H., Weng, T., Chen, P., Hsieh, C., Daniel, L.: Efficient neural network
  robustness certification with general activation functions. In: Neural
  Information Processing Systems (NIPS) (2018)

\bibitem{ZhongTLZK21}
Zhong, Y., Ta, Q., Luo, T., Zhang, F., Khoo, S.: Scalable and modular
  robustness analysis of deep neural networks. In: Proc. of Asian Symposium on
  Programming Languages and Systems (APLAS) (2021)

\end{thebibliography}

\clearpage
\appendix
\section{Offline Proof Sharing}
\label{sec:offline} %

In contrast to the online setting in the main paper, here we discuss an offline setting where we generate templates offline on a training dataset $T_{\text{train}}$ and use them to speed up the verification process on an unobserved data from a test set $T_{\text{test}}$.
Therefore, we consider a modified version of \cref{prob:generation} that, instead of  \cref{eq:generation-opt}, optimizes 
\begin{align}
	&\argmax_{\mathcal{T}} \E_{\vx \sim \mathcal{D}} \left[ \bigvee\limits_{T \in \mathcal{T}} N_{1:k}(\spec(\vx)) \subseteq T  \right]  \label{eq:optoffline}\\
&\text{s.t.}\;\forall \; T \in \mathcal{T}. \, N_{k+1:L}(T) \vdash \post \nonumber,
\end{align}
now with a fixed $\spec$ but $\vx \sim \mathcal{D}$, drawn from the data distribution $\mathcal{D}$.  As standard in machine learning, we assume that the training and test sets, $T_{\text{train}}$ and $T_{\text{test}}$, are sampled from $\mathcal{D}$. An important challenge in the offline setting is that we require the set of templates
$\mathcal{T}$ to be general enough to generalize to unseen inputs from the test set. To this end, in \cref{sec:template-generation-dataset} we describe offline-specific template generation algorithm for solving the optimization in \cref{eq:optoffline}. We note that once we obtain the set of templates $\mathcal{T}$, we use the same procedure as in \cref{alg:verif_online} for utilizing the templates to speed up verification.

\subsection{Template Generation on Training Data}\label{sec:template-generation-dataset} %

We first outline the template generation process in general and then show an
instantiation for $\ell_{\infty}$-robustness verification in \cref{sec:ell_infty-robustness,sec:half-space}.

\subsubsection{Template Generation}

In order to optimize \cref{eq:optoffline} under our
constraints, we attempt to find the set of templates $\mathcal{T}$ with $|\mathcal{T}| \leq m$ that
contains the most abstraction at layer $k$ obtained from
$T_{\text{train}}$. While this is generally computationally hard, we
approximate this with a clustering-based approach.
To this end, we first compute the abstractions 
$N_{1:k}(\spec(\vx))$ for all $\vx \in T_{\text{train}}$ and then
cluster and merge them.
Subsequently, we further merge the obtained templates until we obtain a
set $\mathcal{T}$ of $m$ templates.
We formalize the template generation procedure in
\cref{alg:template-generation-dataset}
and showcase it in \cref{fig:template-generation}.

Similarly to the online template generation, discussed in \cref{sec:template-generation-individual-generation}, we rely on two different verifiers --- the original verifier used to verify inputs $V_S$, and the verifier used to verify our templates $V_T$. We can choose $V_T$ to be more precise and but slower than $V_S$ as the run-time of $V_T$ does not impact the runtime of the inference procedure. We first (\cref{alg:line:proofs},\cref{fig:template-generation-a}) compute the set $\mathcal{V}$ of abstraction computed by $V_S$ at layer $k$, that can be verified. In theory any other verifier could be used for this.
Next, we cluster the abstractions in $\mathcal{V}$ into $n$
groups $\{\mathcal{G}_{i}\mid i = 1,\dots,n\}$ of similar abstractions using the function
\textsc{cluster\_shapes} (which we will instantiate in
\cref{sec:ell_infty-robustness}), showcased by different colors in
\cref{fig:template-generation-b}.
For each group $\mathcal{G}_{i}$, we compute its convex hull $T_i$ via the join operator $\bigsqcup_D$ (\cref{alg:line:join}). The join returns a (usually the tightest) expressible shape in the domain $D$ that includes all its inputs. If we are able to show the post-condition $\post$ for 
$T_i$, then we add the tuple $(T_i, \mathcal{G}_{i})$, the template along
with all abstractions it covers, to the set $\mathcal{H}$,
shown in \cref{fig:template-generation-c} (\cref{alg:line:addpair}).

As depicted in \cref{fig:template-generation-d}, we then attempt to merge
pairs of these templates.  To this end, we traverse the template pairs according to a priority queue $Q$ ordered by a chosen distance $d$ between them (\cref{alg:line:for}). Here, we use the
Euclidean distance between the centers of $T_{i}$ and $T_{j}$ for
$d(T_{i}, T_{j})$. In order to merge $(T_{i}, \mathcal{G}_{i})$ and
$(T_{j}, \mathcal{G}_{j})$, we first compute the set of shapes $\mathcal{G}' = \mathcal{G}_{i} \cup \mathcal{G}_{j}$ contained in either of them
(\cref{alg:line:unionG}) and then again compute the join 
$T' = \bigsqcup_D(\mathcal{G}')$ (\cref{alg:line:union2}). We compue the join this way, as this is likely to result in a tigher overall shape than joining $T_i$ and $T_j$. If $T'$ can be verified, we replace $(T_{i}, \mathcal{G}_{i})$ and
$(T_{j}, \mathcal{G}_{j})$ with the single pair $(T', \mathcal{G}')$ in $\mathcal{H}$ and we update $Q$ accordingly
(\cref{alg:line:mergecheck}~to~\ref{alg:line:mergecheck-done}). This
procedure is repeated until no templates can be merged.
Finally, $\mathcal{T}$ is obtained as the set of the $m$ templates with
the most associated abstractions (e.g., the largest
$|\mathcal{G}|$) in $\mathcal{H}$.

 \begin{algorithm}[ht]
\DontPrintSemicolon
\SetAlgoLined
\SetNoFillComment
\SetKwInOut{KwInput}{Input}
\KwInput{layer number $k$, training set $T_{\text{train}}$, specification $\spec$, verifiers $V_S$, $V_T$}
\KwResult{Set $\mathcal{T}$ of templates, $|\mathcal{T}| \leq m$}
	 $\mathcal{V} \gets \{V_S(\spec(\vx), N_{1:k}) \mid \vx \in T_{\text{train}}, N(\spec(\vx)) \vdash \psi \}$\; \label{alg:line:proofs}
$\mathcal{G}_{1}, \dots, \mathcal{G}_{n} \gets$ \textsc{cluster\_shapes}$(\mathcal{V})$\;
$\mathcal{H} \gets \emptyset$\;
\For{$i \gets 1$ \KwTo $n$}{ \label{alg:line:t}
  $T_i \gets \bigsqcup_{D}(\mathcal{G}_{i})$\; \label{alg:line:union1}\label{alg:line:join}
 \If{$V_T(T_i, N_{k+1:L}) \vdash \psi$}{ \label{alg:line:clustercheck}
    $\mathcal{H} \gets \mathcal{H} \cup \{ (T_i, \mathcal{G}_{i}) \}$\label{alg:line:addpair}\;
  }
}
 $\mathcal{P} \gets \{((T_{i}, \mathcal{G}_{i}), (T_{j}, \mathcal{G}_{j})) \mid  i\neq j,\; (T_{i}, \mathcal{G}_{i}) \in \mathcal{H}\text{, and }(T_{j}, \mathcal{G}_{j})\in\mathcal{H}\}$\; \label{alg:line:mergestart}
 $Q \gets $ priority queue over the pairs in $\mathcal{P}$, ordered by $d(T_{i}, T_{j})$\;
\ForEach{pair $((T_{i}, \mathcal{G}_{i})$, $(T_{j}, \mathcal{G}_{j}))$ in $Q$}{ \label{alg:line:for}
   $\mathcal{G}' = \mathcal{G}_{i} \cup \mathcal{G}_{j}$\; \label{alg:line:unionG}
   $T' = \bigsqcup_{D}(\mathcal{G}')$\; \label{alg:line:union2}
	 \If{$(V_T(T', N_{k+1:L}) \vdash \psi$}{ \label{alg:line:mergecheck}
     $\mathcal{H} \gets \mathcal{H} \setminus \{ (T_{i}, \mathcal{G}_{i}), (T_{j}, \mathcal{G}_{j}) \}$\;
     remove elements containing $(T_{i}, \mathcal{G}_{i})$ from $Q$\;
     remove elements containing $(T_{j}, \mathcal{G}_{j})$ from $Q$\;
     $\mathcal{P}' \gets \{((T, \mathcal{G}), (T', \mathcal{G}')) \mid (T, \mathcal{G}) \in \mathcal{H}\}$\;
     add all pairs from $\mathcal{P}'$ to $Q$\;
     $\mathcal{H} \gets \mathcal{H} \cup \{ (T', \mathcal{G}') \}$\;    
   } \label{alg:line:mergecheck-done}
 }
$\mathcal{T} \gets$ $T$ for $(T, \mathcal{G}) \in \mathcal{H}$ for the $m$ largest $|\mathcal{G}|$\;
\Return{$\mathcal{T}$}\; 
\caption{Template generation over $T_{\text{train}}$}
\label{alg:template-generation-dataset}
\end{algorithm}

\begin{figure}[ht]
    \centering

\def\relaxationsA{
    0.5/0.5/12/8/45,
    1.0/0.7/18/10/75,
    0.9/1.0/18/5/-45
}
\def\relaxationsB{
    1.0/2.5/12/8/105,
    0.5/1.9/16/6/-5,
    1.5/1.8/16/6/60
}
\def\relaxationsC{
    4.5/2.5/12/8/45,
    3.5/2.3/22/6/70,
    4.0/2.1/12/10/0,
    3.5/2.0/12/8/15
}
\def\relaxationsD{
    2.6/0.5/14/12/-25,
    3.6/1.0/18/8/85,
    4.6/0.8/12/8/5
}
\def\relaxationsE{
    2.2/2.4/14/8/15,
    3.8/0.3/14/5/75
}

\def\unionA{
    0.14164/0.14164/1.61770/1.46419
}
\def\unionB{
    0.28432/1.33960/1.99820/2.79265
}
\def\unionC{
    2.76997/1.58624/4.85836/2.85836
}
\def\unionD{
    2.16493/0.01984/4.88245/1.28542
}
\def\unionAB{
    0.14164/0.14164/1.99820/2.79265
}

\subcaptionbox{\label{fig:template-generation-a}}{
  \begin{adjustbox}{width=0.4\textwidth}
\begin{tikzpicture}[font=\small\sffamily]
\fill[my-full-gray] (0, 0) rectangle (5, 3.2);

\foreach \x/\y/\width/\height/\rot in \relaxationsA
{
    \filldraw [my-full-green, rotate around={\rot:(\x,\y)}, opacity=0.2] (\x,\y) ellipse (\height pt and \width pt);
    \draw[my-full-green, rotate around={\rot:(\x,\y)}, opacity=0.5] (\x,\y) ellipse (\height pt and \width pt);
    \draw[fill=my-full-green, opacity=0.5] (\x,\y) circle (0.02);
}
\foreach \x/\y/\width/\height/\rot in \relaxationsB
{
    \filldraw [my-full-green, rotate around={\rot:(\x,\y)}, opacity=0.2] (\x,\y) ellipse (\height pt and \width pt);
    \draw[my-full-green, rotate around={\rot:(\x,\y)}, opacity=0.5] (\x,\y) ellipse (\height pt and \width pt);
    \draw[fill=my-full-green, opacity=0.5] (\x,\y) circle (0.02);
}
\foreach \x/\y/\width/\height/\rot in \relaxationsC
{
    \filldraw [my-full-green, rotate around={\rot:(\x,\y)}, opacity=0.2] (\x,\y) ellipse (\height pt and \width pt);
    \draw[my-full-green, rotate around={\rot:(\x,\y)}, opacity=0.5] (\x,\y) ellipse (\height pt and \width pt);
    \draw[fill=my-full-green, opacity=0.5] (\x,\y) circle (0.02);
}
\foreach \x/\y/\width/\height/\rot in \relaxationsD
{
    \filldraw [my-full-green, rotate around={\rot:(\x,\y)}, opacity=0.2] (\x,\y) ellipse (\height pt and \width pt);
    \draw[my-full-green, rotate around={\rot:(\x,\y)}, opacity=0.5] (\x,\y) ellipse (\height pt and \width pt);
    \draw[fill=my-full-green, opacity=0.5] (\x,\y) circle (0.02);
}
\foreach \x/\y/\width/\height/\rot in \relaxationsE
{
    \filldraw [my-full-red, rotate around={\rot:(\x,\y)}, opacity=0.2] (\x,\y) ellipse (\height pt and \width pt);
    \draw[my-full-red, rotate around={\rot:(\x,\y)}, opacity=0.5] (\x,\y) ellipse (\height pt and \width pt);
    \draw[fill=my-full-red, opacity=0.5] (\x,\y) circle (0.02);
}

\end{tikzpicture}
\end{adjustbox}
}
\hfill
\subcaptionbox{\label{fig:template-generation-b}}{
  \begin{adjustbox}{width=0.4\textwidth}
\begin{tikzpicture}[font=\small\sffamily]

\fill[my-full-gray] (0, 0) rectangle (5, 3.2);

\foreach \x/\y/\width/\height/\rot in \relaxationsA
{
    \filldraw [my-full-blue, rotate around={\rot:(\x,\y)}, opacity=0.2] (\x,\y) ellipse (\height pt and \width pt);
    \draw[my-full-blue, rotate around={\rot:(\x,\y)}, opacity=0.5] (\x,\y) ellipse (\height pt and \width pt);
    \draw[fill=my-full-blue, opacity=0.5] (\x,\y) circle (0.02);
}
\foreach \x/\y/\width/\height/\rot in \relaxationsB
{
    \filldraw [my-full-orange, rotate around={\rot:(\x,\y)}, opacity=0.2] (\x,\y) ellipse (\height pt and \width pt);
    \draw[my-full-orange, rotate around={\rot:(\x,\y)}, opacity=0.5] (\x,\y) ellipse (\height pt and \width pt);
    \draw[fill=my-full-orange, opacity=0.5] (\x,\y) circle (0.02);
}
\foreach \x/\y/\width/\height/\rot in \relaxationsC
{
    \filldraw [my-full-cyan, rotate around={\rot:(\x,\y)}, opacity=0.2] (\x,\y) ellipse (\height pt and \width pt);
    \draw[my-full-cyan, rotate around={\rot:(\x,\y)}, opacity=0.5] (\x,\y) ellipse (\height pt and \width pt);
    \draw[fill=my-full-cyan, opacity=0.5] (\x,\y) circle (0.02);
}
\foreach \x/\y/\width/\height/\rot in \relaxationsD
{
    \filldraw [my-full-olive, rotate around={\rot:(\x,\y)}, opacity=0.2] (\x,\y) ellipse (\height pt and \width pt);
    \draw[my-full-olive, rotate around={\rot:(\x,\y)}, opacity=0.5] (\x,\y) ellipse (\height pt and \width pt);
    \draw[fill=my-full-olive, opacity=0.5] (\x,\y) circle (0.02);
}

\foreach \xlb/\ylb/\xub/\yub in \unionA
{
    \filldraw[my-full-blue, opacity=0.5, fill opacity=0.1] (\xlb, \ylb) rectangle (\xub, \yub);
}
\foreach \xlb/\ylb/\xub/\yub in \unionB
{
    \filldraw[my-full-orange, opacity=0.5, fill opacity=0.1] (\xlb, \ylb) rectangle (\xub, \yub);
}
\foreach \xlb/\ylb/\xub/\yub in \unionC
{
    \filldraw[my-full-cyan, opacity=0.5, fill opacity=0.1] (\xlb, \ylb) rectangle (\xub, \yub);
}
\foreach \xlb/\ylb/\xub/\yub in \unionD
{
    \filldraw[my-full-olive, opacity=0.5, fill opacity=0.1] (\xlb, \ylb) rectangle (\xub, \yub);
}

\end{tikzpicture}
\end{adjustbox}
}\\
\subcaptionbox{\label{fig:template-generation-c}}{
  \begin{adjustbox}{width=0.4\textwidth}
\begin{tikzpicture}[font=\small\sffamily]
\fill[my-full-gray] (0, 0) rectangle (5, 3.2);

\foreach \x/\y/\width/\height/\rot in \relaxationsA
{
    \filldraw [my-full-green, rotate around={\rot:(\x,\y)}, opacity=0.2] (\x,\y) ellipse (\height pt and \width pt);
    \draw[my-full-green, rotate around={\rot:(\x,\y)}, opacity=0.5] (\x,\y) ellipse (\height pt and \width pt);
    \draw[fill=my-full-green, opacity=0.5] (\x,\y) circle (0.02);
}
\foreach \x/\y/\width/\height/\rot in \relaxationsB
{
    \filldraw [my-full-green, rotate around={\rot:(\x,\y)}, opacity=0.2] (\x,\y) ellipse (\height pt and \width pt);
    \draw[my-full-green, rotate around={\rot:(\x,\y)}, opacity=0.5] (\x,\y) ellipse (\height pt and \width pt);
    \draw[fill=my-full-green, opacity=0.5] (\x,\y) circle (0.02);
}
\foreach \x/\y/\width/\height/\rot in \relaxationsC
{
    \filldraw [my-full-green, rotate around={\rot:(\x,\y)}, opacity=0.2] (\x,\y) ellipse (\height pt and \width pt);
    \draw[my-full-green, rotate around={\rot:(\x,\y)}, opacity=0.5] (\x,\y) ellipse (\height pt and \width pt);
    \draw[fill=my-full-green, opacity=0.5] (\x,\y) circle (0.02);
}
\foreach \x/\y/\width/\height/\rot in \relaxationsD
{
    \filldraw [my-full-red, rotate around={\rot:(\x,\y)}, opacity=0.2] (\x,\y) ellipse (\height pt and \width pt);
    \draw[my-full-red, rotate around={\rot:(\x,\y)}, opacity=0.5] (\x,\y) ellipse (\height pt and \width pt);
    \draw[fill=my-full-red, opacity=0.5] (\x,\y) circle (0.02);
}

\foreach \xlb/\ylb/\xub/\yub in \unionA
{
    \filldraw[my-full-green, opacity=0.5, fill opacity=0.1] (\xlb, \ylb) rectangle (\xub, \yub);
}
\foreach \xlb/\ylb/\xub/\yub in \unionB
{
    \filldraw[my-full-green, opacity=0.5, fill opacity=0.1] (\xlb, \ylb) rectangle (\xub, \yub);
}
\foreach \xlb/\ylb/\xub/\yub in \unionC
{
    \filldraw[my-full-green, opacity=0.5, fill opacity=0.1] (\xlb, \ylb) rectangle (\xub, \yub);
}
\foreach \xlb/\ylb/\xub/\yub in \unionD
{
    \filldraw[my-full-red, opacity=0.5, fill opacity=0.1] (\xlb, \ylb) rectangle (\xub, \yub);
}

\end{tikzpicture}
\end{adjustbox}
}
\hfill
\subcaptionbox{\label{fig:template-generation-d}}{
  \begin{adjustbox}{width=0.4\textwidth}
\begin{tikzpicture}[font=\small\sffamily]
\fill[my-full-gray] (0, 0) rectangle (5, 3.2);

\foreach \x/\y/\width/\height/\rot in \relaxationsA
{
    \filldraw [my-full-green, rotate around={\rot:(\x,\y)}, opacity=0.2] (\x,\y) ellipse (\height pt and \width pt);
    \draw[my-full-green, rotate around={\rot:(\x,\y)}, opacity=0.5] (\x,\y) ellipse (\height pt and \width pt);
    \draw[fill=my-full-green, opacity=0.5] (\x,\y) circle (0.02);
}
\foreach \x/\y/\width/\height/\rot in \relaxationsB
{
    \filldraw [my-full-green, rotate around={\rot:(\x,\y)}, opacity=0.2] (\x,\y) ellipse (\height pt and \width pt);
    \draw[my-full-green, rotate around={\rot:(\x,\y)}, opacity=0.5] (\x,\y) ellipse (\height pt and \width pt);
    \draw[fill=my-full-green, opacity=0.5] (\x,\y) circle (0.02);
}
\foreach \x/\y/\width/\height/\rot in \relaxationsC
{
    \filldraw [my-full-green, rotate around={\rot:(\x,\y)}, opacity=0.2] (\x,\y) ellipse (\height pt and \width pt);
    \draw[my-full-green, rotate around={\rot:(\x,\y)}, opacity=0.5] (\x,\y) ellipse (\height pt and \width pt);
    \draw[fill=my-full-green, opacity=0.5] (\x,\y) circle (0.02);
}

\foreach \xlb/\ylb/\xub/\yub in \unionA
{
    \draw[my-full-green, opacity=0.5, dashed] (\xlb, \ylb) rectangle (\xub, \yub);
}
\foreach \xlb/\ylb/\xub/\yub in \unionB
{
    \draw[my-full-green, opacity=0.5, dashed] (\xlb, \ylb) rectangle (\xub, \yub);
}
\foreach \xlb/\ylb/\xub/\yub in \unionC
{
    \filldraw[my-full-green, opacity=0.5, fill opacity=0.1] (\xlb, \ylb) rectangle (\xub, \yub);
}
\foreach \xlb/\ylb/\xub/\yub in \unionAB
{
    \filldraw[my-full-green, opacity=0.5, fill opacity=0.1] (\xlb, \ylb) rectangle (\xub, \yub);
}

\end{tikzpicture}
\end{adjustbox}
}
\caption{Visualization of \cref{alg:template-generation-dataset}.
First, in (a) the abstractions at layer $k$ for input regions in the training set are obtained, and restricted to the verifiable ones (green).
These are then clustered and their convex hulls in domain $D$ are obtained. (b) shows different clusters in different colors.
The convex hulls are then verified (c), and restricted to the verifiable ones (green).
Finally in (d), these regions are further merged, if possible, to obtain the set of templates.}
\label{fig:template-generation}
\end{figure}

\subsection{Dataset templates for $\ell_\infty$ robustness}\label{sec:ell_infty-robustness} %
We now instantiate the template generation algorithm in \cref{sec:template-generation-dataset} for speeding up
$\ell_{\infty}$-robustness verification $\spec_{\epsilon}$.  As in
\cref{sec:template-generation-individual}, we rely on verifier $V_S$ based on Zonotopes and represent the templates as Boxes (with possible additional half-space constraints as outlined below).
For the verification of the templates
(lines~\ref{alg:line:clustercheck}~and~\ref{alg:line:mergecheck})
we perform exact verification via Mixed-Integer
Linear Programming (MILP) \cite{TjengMILP} via verifier $V_T$.
The box-encoded templates can be directly verified by the exact
verifier. We note that since exact verification is strictly more precise
than Zonotope propagation, the use of templates can potentially allow for
\emph{higher} certification rates than directly employing Zonotope propagation.
While we did not observe this experimentally, it presents an
interesting target for further investigation.

We instantiate the join $\bigsqcup_D$ with the join in the Box domain $\bigsqcup_{\text{B}}$.
For a set of Zonotopes $\mathcal{G} = \{\mathcal{V}_1, \dots, \mathcal{V}_n\}$, we compute the bounding box $\alpha_{\text{Box}}(\mathcal{V}_i)$ for all Zonotopes and then compute the joined bounding Box (which again can be represented as a Zonotope).

\subsubsection{Exact verification} We now briefly outline the properties of exact verification via MILP, as we require these
in the following discussion. The framework from \cite{TjengMILP},
proves classification to the correct label $l$ by maximizing the error term
$e = \max_{i \neq l} \vn_{i} - \vn_{l}$ and asserting that $e < 0$, where $\vn$
denotes the output of the neural network (e.g. its logits) over the considered
input region.
If no counterexample to that assertion can be found, it certifies the
specification, else it returns a set of counterexamples
$\{\vz_{V,i}\}$ (concrete points in the input region), utilized later, for which this error is maximal. In
both cases we can access value $e$ of the error function.

\subsubsection{Shape clustering} Next, we describe how we instantiate the clustering method \textsc{cluster\_shapes} in this setting. We base \textsc{cluster\_shapes} on $k$-means clustering for which we provide a similarity matrix computed as follows. For each
pair of Zonotopes in $\{(\mathcal{V}_{i}, \mathcal{V}_{j}) \mid \mathcal{V}_{i},\mathcal{V}_{j} \in \mathcal{V}\}$, we compute the joined Box
$B_{i,j} = \alpha_{\text{Box}}(\mathcal{V}_{i}) \sqcup_{B} \alpha_{\text{Box}}(\mathcal{V}_{j})$, where $\sqcup_{B}$ denotes the Box join operator.
We then set the distance between $\mathcal{V}_{i}$ and $\mathcal{V}_{j}$ to $\exp(e)$, where $e \in \mathbb{R}$ is the error obtained from exact verification when attempting to verify $\psi$ for $B_{i,j}$.
To obtain a similarity matrix from these distances, we apply a
constant shift embedding \cite{RothLKB03}. As invoking exact verification on
each box $B_{i,j}$ is expensive, we only consider the $t$
closest neighbors (in $\ell_{2}$ distance between the Zonotope centers) and set
all others to a maximal distance.

\subsubsection{Half-space constraints}\label{sec:half-space} %
To allow for templates $T$ covering more volume, e.g., those that allow to optimize
\cref{eq:optoffline} further by containing more abstractions,
we extend the template representation from Boxes to Boxes with additional half-space
constraints, formally called Stars \cite{tran2019star,bak2017simulation}.
As the Star domain is more precise than the Box domain (by allowing to cut some of the box volume), using Stars enables us to generate templates with higher volume  that are still verifiable by $V_T$. Further, the Star domain allows efficient containment checks $S \subseteq T$ similarly to the Box domain. Formally a Star $B^{*}$ over a Box $B$ is denoted as:

\begin{align}
  &B^{*}(\mC, \vc) := \{\vz \in \mathbb{R}^{d} \mid \vz \in B \wedge C \vz \leq \vc \} \label{eq:star}\\
  &\text{with} \; \mC \in  \mathbb{R}^{c \times d}, \vc \in \mathbb{R}^{c}. \nonumber
\end{align}

Here each half-space constraint is described by a hyperplane parameterized
by $\mC_{i, \cdot}$ and $c_{i}$.

The containment check $S \subseteq B^{*}(\mC, \vc)$ between an abstraction $S$ and the Star $B^{*}(\mC, \vc)$ consists of: (i) a containment for the
underlying box $S \subseteq B$, and (ii) checking if for each constraint
$C_i \cdot z \leq c_i$, maximizing the linear expression $C_i \cdot z$
with respect to $S$ yields an objective $\leq c_i$.
For a Zonotope $S$ as given in \cref{eq:zonotope}, in \cref{sec:domains} we showed how to perform step (i) efficiently and step (ii) can be performed efficiently by checking the condition $\mC\va + \sum_{j=1}^p |\mC\mA|_{j} \leq \vc$.

A star encoded as in \cref{eq:star} can be directly verified using
exact verification (MILP) by adding the half-space constraints as further LP
constraints.

\subsubsection{Obtaining half-space contraints} %
In the template generation process we utilize Boxes as
before. However, whenever we fail to verify a template (e.g.,
lines~\ref{alg:line:clustercheck}~and~\ref{alg:line:mergecheck} in
\cref{alg:template-generation-dataset}), we attempt to add a
half-space constraint. We repeat this up to $n_{\text{hs}}$ times
resulting in as many constraints.
We leverage the exact verifier for obtaining half-space
constraints.  Recall, that it either verifies a region or provides a set of
counterexamples $\{\vz_{V,i}\}$.
Since we only add additional half-space constraints, if the verification
fails we utilize these counterexamples. In the following we assume
a single $\vz_{V}$, and derive a hyperplane that separates $\vz_{V}$ from the abstraction we are
trying to verify.  If there are multiple $\vz_{V,i}$, we iterate over them and
perform the described procedure for each $\vz_{V,i}$, that is not already cut
by the hyperplane found for a previous counterexample.
These hyperplanes directly yield the new constraints.

We showcase this in \cref{fig:hyperplanes}, where the green shaded
area $T$ shows the Box join over three abstraction 
$T = \bigsqcup_{B}(\{P_{1}, P_{2}, P_{3}\})$ The individual $P_{i}$,
shown in blue, are zonotopes, that can be verified individually.

The verification of the green area fails, with the counterexample
$\vz_{V}$ (red dot) shown in the top right corner. To find a
hyperplane that separates $\vz_{V}$ from the rest of $T$, we consider
the line from the center $\va$ (green dot) of $T$ to the point
$\vz_{V}$ and a hyperplane orthogonal to it (shown as the dashed
line). Thus, adding a row $i$ to the matrix $\mC$ of the star:
$\mC_i = (\vz_{V} - \va)^{T}$.  To find offset
$\vh = \lambda \va + (1-\lambda) \vz_{V}$ (for $\lambda \in [0, 1]$)
along this line to, we consider the value
attained for $\mC_i \vx$ for $\vx$ in the verified area
($P_{1}, P_{2}, P_{3}$):
\begin{equation*}
  c_p := max_{\substack{\vx \in P\\P \in \{P_{1}, P_{2}, P_{3}\}}} \mC_i\vx.
\end{equation*}
Then, the constant $\vc_{i}$ of the new hyperplane is
given by $\vc_i = \kappa \vc_p + (1-\kappa) \vc_{z}$ for a
hyper-parameter $\kappa$ and $\vc_z := \mC_i\vz_V$.

A high $\kappa$ puts the hyperplane close to $\vz_V$, removes only
little volume from the template, while low $\kappa$ puts it closer to
$\va$.  Since $\vz_V$ is only the counterexample with the largest
violation, but not necessarily the whole region preventing
certification, the half-space constraint obtained from a high $\kappa$
might not be sufficient to separate this region from $T$. Thus, in a
subsequent iteration, another half-space constraint for the same region
may be added.  For low $\kappa$, fewer constraints are required, but
more verifiable volume of $T$ is lost.

\begin{figure}[t]
  \centering
  \begin{adjustbox}{width=0.350\textwidth}
    \begin{tikzpicture}
\fill[my-full-gray] (0, 0) rectangle (5, 3.2);
\fill[my-full-green, opacity=0.05, line width=0.8pt] (0.2, 0.2) rectangle (4.8, 2.8);
\fill[my-full-green, opacity=0.1]
(0.2, 0.2) -- (4.8, 0.2) -- (4.8,1.189) -- (3.890,2.8) -- (0.2,2.8) -- cycle;
\draw[my-full-blue, opacity=0.5, line width=0.6pt]
(0.2, 0.2) --
(1.2, 0.2) --
(1.5, 1.2) --
(0.5, 1.2) -- cycle;
\draw[my-full-blue, opacity=0.5, line width=0.6pt]
(3.0, 0.2) --
(4.8, 0.2) --
(4.0, 1.0) --
(2.2, 1.0) -- cycle;
\draw[my-full-blue, opacity=0.5, line width=0.6pt]
(0.7, 2.3) --
(2.2, 2.3) --
(2.5, 2.8) --
(1.0, 2.8) -- cycle;

\coordinate (center) at (2.5, 1.5);
\coordinate (counterexample) at (4.8, 2.8);
\draw[fill=my-full-green, draw=my-full-green] (center) circle (0.05);
\draw[fill=my-full-red, draw=my-full-red] (counterexample) circle (0.05);
\draw[black, ->] (center) -- (counterexample);
\node (c) [above=0.2mm of center] {\tiny $\va$};
\node (cE) [above=0.2mm of counterexample] {\tiny $\vz_{V}$};

\node[text=my-full-green, opacity=0.8] at (0.4, 2.6) {\tiny $T$};
\node[text=my-full-blue, opacity=0.8] at (0.85, 0.7) {\tiny $P_{1}$};
\node[text=my-full-blue, opacity=0.8] at (3.5, 0.6) {\tiny $P_{2}$};
\node[text=my-full-blue, opacity=0.8] at (1.6, 2.55) {\tiny $P_{3}$};

\coordinate (p) at ($(center)!0.7!(counterexample)$);
\coordinate (direction) at ($(counterexample)-(center)$);
\path let \p1 = (direction) in coordinate (directionT) at (-\y1,\x1);

\coordinate (p1) at ($(p)+(directionT)$);
\coordinate (p2) at ($(p)-(directionT)$);
\draw[my-full-green, line width=0.8pt, dashed] ($(p)!0.2!(p1)$) -- ($(p)!0.6!(p2)$);
\draw[fill=my-full-blue, draw=my-full-blue] (p) circle (0.05);
\node (pE) [above=0.2mm of p] {\tiny $\vh$};
\node[] at (3.5, 2.3) {\tiny $\mC_{i}$};
\node[] at (4.4, 1.5) {\tiny $\vc_{i}$};

\end{tikzpicture}
\end{adjustbox}
\caption{The algorithm used to find half-space constraints, by cutting
  counterexample $\vz_{V}$ from the template $T$ with a hyperplane
  ($\mC_{i}, \vc_{i}$).  The normal of the hyperplane (specified by
  $\mC_i$) is given by the vector between $\va$ (the center of $T$)
  and $\vz_V$.  The threshold $\vc_i$ is chosen such that the
  hyperplane remove $\vz_V$ but does not intersect any relaxations $P_{1}, P_{2}, P_{3}$ the template $T$ was created from.}
\label{fig:hyperplanes}
\end{figure} %

\subsection{Template Expansion}\label{sec:template-extension} %
\begin{algorithm}[ht]
\DontPrintSemicolon
\SetAlgoLined
\SetNoFillComment
\SetKwInOut{KwInput}{Input}
\KwInput{layer number $k$, templates $\mathcal{T} = \{T_{i}\}_{i=1}^{m}$ at layer $k$, scaling matrix $\mD$, verifier $V_T$}
\KwResult{Set $\mathcal{T}^w$ of expanded templates, $|\mathcal{T}^w| = m$}
\For{$i \gets 1$ \KwTo $m$}{
  $T^{w}_i \gets T_{i}$\;
  $T_{i} \gets \mD \cdot T_{i}$\;
  \While{$V_T(T_i, N_{k+1:L}) \vdash \psi$}{ \label{alg:line:wideningcheck}
    $T^{w}_i \gets T_{i}$\;
    \If{$T_i$ is Star}{
    	$T_{i} \gets$ \textsc{remove\_planes}$(T_{i})$\;\label{alg:line:widenplanes}
    }
    $T_{i} \gets \mD \cdot T_{i}$\;\label{alg:line:scale_wide}
    \If{$T_i$ is Star}{
    	$T_{i} \gets$ \textsc{copy\_planes}$(T^{w}_{i})$\;\label{alg:line:widenplanes}
    \If{$V_T(T_i, N_{k+1:L}) \not\vdash \psi$}{
    	$T_{i} \gets$ \textsc{add\_planes}$(T_{i}, n^\text{te}_\text{hs})$\;\label{alg:line:widenplanesnew}
    }
	}
  }
}
\Return{$\mathcal{T}^w = \{T^w_{i}\}_{i=1}^{m}$}
\caption{Template expansion}
\label{alg:template-widening}
\end{algorithm}

To further improve the generalization of our templates from the training set to the test set,
we introduce an operation called \emph{template expansion}, outlined in
\cref{alg:template-widening}. We apply the template expansion to the result of the template generation presented in \cref{alg:template-generation-dataset} and we use the resulting widened templates for our offline proof transfer algorithm. \cref{alg:template-widening} tries to expand each of the templates $T_i$ separately, by repeatedly scaling the template's associated Box by a diagonal scaling matrix $\mD := \diag(f_{1}, \dots, f_{d})$ (\cref{alg:line:scale_wide}) until $T_i$ is no longer verifiable by the template verifier $V_T$ (\cref{alg:line:wideningcheck}). Here, $f_{j} \geq 1$ acts as a scaling factor for the $j$-th dimension of the template Boxes.

\subsubsection{Template Expansion for Stars} %
When we encode templates as Stars,
the scaling matrix is only applied on the Star's underlying Box,
but not on the half-space constraints, since these have already been selected to be close to the decision boundary. Thus for the new extened template $T_i$ we copy the constraints from $T_i^w$ (\cref{alg:line:widenplanes}).
If the resulting template fails to verify, we generate up to $n^\text{te}_\text{hs}$ additional constraints, in the same way as for \cref{alg:template-generation-dataset}, and add them to $T_i$ (\cref{alg:line:widenplanesnew}).

\begin{table*}
    \centering
    \begin{small}

    \begin{small}

      \caption{
        Template matching rate and verification time of the
        whole MNIST test set $t$ in seconds for the 5x100 using up to
        $m$ templates per label and layer pair. The baseline verification $292.13 \pm 1.77$ and
        $291.80 \pm 2.36$ seconds for $\epsilon=0.05$ and $\epsilon=0.10$ respectively. +TE indicates the use of Template Expansion. }
    \label{tab:l_infinity_5x100_full}

	    \vspace{2mm}

    \begin{tabular}{@{}rcrrrcrrr@{}}\toprule
    \textbf{Box}& &\multicolumn{3}{c}{shapes matched $[\%]$}&&\multicolumn{3}{c}{verification time [s]}\\
    \cmidrule{3-5}\cmidrule{7-9}
    $k$ && $m=1$ & $m=3$ & $m=25$ && $m=1$ & $m=3$ & $m=25$\\
   \midrule
    $\epsilon=0.05$\\
    3 && 07.5 & 14.4 & 28.4 && $282.60 \pm 2.18$ & $275.32\pm 1.34$ & $259.87 \pm 0.68$\\
      4 && 19.5 & 38.0 & 57.6 && $279.87 \pm 0.07$ & $270.08 \pm 0.81$ & $258.06 \pm 1.23$\\
    3+4 && 20.7 & 39.3 & 59.2 && $274.86 \pm 0.31$ & $259.19 \pm 0.91$ & $243.11 \pm 0.81$\\
   \midrule
    $\epsilon=0.10$\\
    3 && 05.1 & 10.2 & 19.4 && $286.67 \pm 1.17$ & $280.45 \pm 1.03$ & $272.01 \pm 0.68$\\
    4 && 15.0 & 29.6 & 45.8 && $283.06 \pm 1.49$ & $275.39 \pm 1.04$ & $268.05 \pm 1.04$\\
    3+4 && 15.8 & 30.7 & 47.4 && $281.14 \pm 1.12$ & $272.17 \pm 1.18$ & $257.92 \pm 0.97$\\
    \bottomrule
    \end{tabular}
    \hfill
    \begin{tabular}{@{}rcrrrcrrr@{}}\toprule
    \textbf{Box+TE}& &\multicolumn{3}{c}{shapes matched $[\%]$}&&\multicolumn{3}{c}{verification time [s]}\\
    \cmidrule{3-5}\cmidrule{7-9}
    $k$ && $m=1$ & $m=3$ & $m=25$ && $m=1$ & $m=3$ & $m=25$\\
   \midrule
    $\epsilon=0.05$\\
    3 && 7.8 & 15.0 & 30.5 && $284.27 \pm 0.63$ & $274.26 \pm 1.78$ & $257.77 \pm 1.21$\\
    4 && 20.1 & 38.9 & 59.0 && $280.44 \pm 1.66$ & $268.97 \pm 1.13$ & $258.02 \pm 1.38$\\
    3+4 && 21.3 & 36.0 & 60.6 && $275.31 \pm 2.62$ & $260.95 \pm 0.70$ & $241.36 \pm 1.34$ \\
    \midrule
    $\epsilon=0.10$\\
    3 && 5.4 & 10.6 & 21.2 && $285.76 \pm 0.71$ & $278.94 \pm 1.12$ & $266.85 \pm 2.07$\\
    4 && 15.6 & 30.5 & 47.7 && $285.15 \pm 0.47$ & $274.82 \pm 1.41$ & $266.62 \pm 0.45$\\
    3+4 && 16.3 & 31.5 & 49.2 && $280.49 \pm 0.70$ & $269.90 \pm 0.34$ & $257.82 \pm 3.01$\\
   \bottomrule
    \end{tabular}\\

    \begin{tabular}{@{}rcrrrcrrr@{}}\toprule
     \textbf{Star}& &\multicolumn{3}{c}{shapes matched  $[\%]$}&&\multicolumn{3}{c}{verification time [s]}\\
     \cmidrule{3-5}\cmidrule{7-9}
     $k$ && $m=1$ & $m=3$ & $m=25$ && $m=1$ & $m=3$ & $m=25$\\
    \midrule
     $\epsilon=0.05$\\
     3 && 10.8 & 25.2 & 40.4 && $281.33 \pm 1.44$ & $269.24 \pm 1.30$ & $254.41 \pm 0.87$\\
     4 && 25.8 & 40.3 & 62.7 && $282.51 \pm 2.10$ & $281.68 \pm 0.60$ & $271.63 \pm 0.89$\\
     3+4 && 27.9 & 45.1 & 64.3 && $277.89 \pm 1.38$ & $266.55 \pm 0.93$ & $246.39 \pm 0.31$\\
    \midrule
    $\epsilon=0.10$\\
    3 && 7.9 & 18.2 & 29.0 && $284.31 \pm 2.19$ & $277.96 \pm 0.60$ & $267.72 \pm 0.85$\\
    4 && 20.2 & 31.9 & 51.0 && $285.68 \pm 2.10$ & $286.16 \pm 0.79$ & $278.29 \pm 1.30$\\
    3+4 && 21.7 & 35.8 & 52.9 && $283.43 \pm 0.91$ & $278.00 \pm 0.60$ & $262.96 \pm 1.04$\\
     \bottomrule
    \end{tabular}
\hfill
   \begin{tabular}{@{}rcrrrcrrr@{}}\toprule
    \textbf{Star+TE}& &\multicolumn{3}{c}{shapes matched $[\%]$}&&\multicolumn{3}{c}{verification time [s]}\\
    \cmidrule{3-5}\cmidrule{7-9}
    $k$ && $m=1$ & $m=3$ & $m=25$ && $m=1$ & $m=3$ & $m=25$\\
   \midrule
    $\epsilon=0.05$\\
    3 && 11.1 & 25.8 & 41.8 && $282.99 \pm 0.45$ & $271.34 \pm 1.89$ & $254.62 \pm 1.05$\\
    4 && 25.9 & 40.6 & 63.6 && $282.98 \pm 1.46$ & $281.51 \pm 0.23$ & $270.44 \pm 0.30$\\
    3+4 && 28.3 & 45.6 & 65.2 && $278.00 \pm 0.36$ & $269.08 \pm 0.80$ & $247.43 \pm 0.09$\\
   \midrule
    $\epsilon=0.10$\\
    3 && 8.1 & 18.7 & 30.3 && $285.82 \pm 1.91$ & $280.71 \pm 2.99$ & $267.55 \pm 0.77$\\
    4 && 20.3 & 32.3 & 52.8 && $285.49 \pm 1.04$ & $286.34 \pm 0.79$ & $276.14 \pm 0.30$\\
    3+4 && 22.0 & 36.2 & 54.5 && $284.39 \pm 2.29$ & $278.90 \pm 0.34$ & $262.60 \pm 1.71$\\
    \bottomrule
    \end{tabular}\\

    \end{small}

\end{small}
\end{table*}

\subsection{Experimental Evaluation}\label{sec:evaluation-linf} %
In this section, we instantiate our offline proof transfer to the MNIST dataset. We consider templates both in the Box and Star domains both with and without template expansion (\cref{sec:template-extension}) and 5x100 fully-connected network with ReLU activations. We generate templates  individually for every label at both the third and fourth layer. For technical details see the end of this section. We allow up to $m=25$ templates for each combination. We experiment with two different values for $\epsilon$: $0.05$  and $0.1$.

\cref{tab:l_infinity_5x100_full} shows the results.
We provide the
fraction of input regions that could be successfully matched to
templates as well as the overall verification time.
If an input cannot be matched with any of the templates, then we
propagate the standard Zonotope abstraction through the rest of the
network to verify it.

We observe that templates can subsume up to 57.6\% of input regions in the test set with
$\epsilon=0.05$, and up to 45.8\% for the higher $\epsilon=0.1$ when expressed in the Box domain (at layer 4).
Additionally enabling template expansion increases these rates to 59.0\% and 47.7\% respectively.
Combining template at multiple layers gives more matched templates as many inputs can be matched in the third layer, while unmatched ones can again be considered at the fourth layer for a total of up to 60.6\% and 49.2\% respecitvely.
We observe that improvements in matching rate directly lead to speed ups over standard verification.
Additionally allowing half-space constraints, i.e., using Stars instead of Boxes as templates, allows us to increase the matching rate up to 65.2 \% and 54.5 \% for the two $\epsilon$ respectively when using TE. However, as checking matches for Stars is computationally more expensive the resulting final verification time is slightly worse compared to Box templates.

To summarize, these results highlight that with the algorithm outlined in
\cref{sec:ell_infty-robustness}, a set of templates $\mathcal{T}$ can
be obtained that generalize remarkably well to new unseen input regions
(e.g., up to 65.2 \% containment).
More precise abstractions such as Stars allow templates that capture a far higher rate of containment for new input regions, the added cost of their containment check makes the obtained speedups smaller.
Finally, we see that Template Expansion (\cref{sec:template-extension}) uniformly leads to a higher matching rate and speed-ups.

\subsubsection{Technical Details}
We use a feed forward neural network with five linear layers of size 100 and ReLU activations, trained with DiffAI \cite{DiffAI}. The network has an
accuracy of $0.94$ and a certified accuracy of $0.93$ and $0.92$ for $\epsilon = 0.1$
and $\epsilon = 0.2$ respectively. 

For a \textsc{cluster\_proofs} we set an initial cluster size depending on the
number of verifiable images per label, in order
that the clusters contain on average 50 images.
For the verification of a cluster's union, we allow up to $n_{\text{hs}} = 30$ half-space constraints
and set $\kappa$ of 0.05. Taking a low value leads to a larger truncation,
but reduces the number of half-space constraints, which speeds up the template generation
as well as the containment check at inference.
We take the same values also for verifying unions after merging two clusters.
For expanding the templates, we use up to 10 iterations, in which we widen by 5\%
in each dimension and then allow up to 10 hyperplanes to verify the expanded template.
To avoid truncating previously verified volume, we increase $\kappa$ linearly by $0.02$
for each expansion step, starting with an initial $\kappa$ of 0.4

\end{document}